\documentclass[preprint,12pt]{elsarticle}

\usepackage[table]{xcolor}
\usepackage{times}
\usepackage{latexsym}
\usepackage[T1]{fontenc}
\usepackage[utf8]{inputenc}
\usepackage{microtype}
\usepackage{inconsolata}

\usepackage{graphicx}
\usepackage{comment}
\usepackage{subcaption}
\usepackage{booktabs}
\usepackage{multirow}
\usepackage{pifont}
\usepackage{array}
\usepackage[mathscr]{eucal}
\usepackage{amsmath}
\usepackage{float}
\usepackage{amssymb}

\usepackage{tabularx}
\usepackage{adjustbox}

\usepackage[normalem]{ulem}
\useunder{\uline}{\ul}{}

\usepackage{color}
\usepackage{tcolorbox}

\journal{Knowledge-Based Systems}

\begin{document}

\begin{frontmatter}

\title{Question Answering Over Spatio-temporal Knowledge Graph}

\author{Xinbang Dai\fnref{label1}}
\author{Huiying Li\fnref{label1}}
\author{Nan Hu\fnref{label1}}
\author{Yongrui Chen\fnref{label1}}
\author{Rihui Jin\fnref{label1}}
\author{Huikang Hu\fnref{label1}}
\author{Guilin Qi\corref{cor1}\fnref{label1}}
\ead{gqi@seu.edu.cn}
\cortext[cor1]{Corresponding author.}
\affiliation[label1]{organization={Southeast University},
            city={Nanjing},
            state={Jiangsu},
            country={China}}

\begin{abstract}
Spatio-temporal knowledge graphs (STKGs) enhance traditional KGs by integrating temporal and spatial annotations, enabling precise reasoning over questions with spatio-temporal dependencies. Despite their potential, research on spatio-temporal knowledge graph question answering (STKGQA) remains limited. This is primarily due to the lack of datasets that simultaneously contain spatio-temporal information, as well as methods capable of handling implicit spatio-temporal reasoning. To bridge this gap, we introduce the spatio-temporal question answering dataset (STQAD), the first comprehensive benchmark comprising 10,000 natural language questions that require both temporal and spatial reasoning. STQAD is constructed with real-world facts containing spatio-temporal information, ensuring that the dataset reflects practical scenarios. Furthermore, our experiments reveal that existing KGQA methods underperform on STQAD, primarily due to their inability to model spatio-temporal interactions. To address this, we propose the spatio-temporal complex question answering (STCQA) method, which jointly embeds temporal and spatial features into KG representations and dynamically filters answers through constraint-aware reasoning. STCQA achieves state-of-the-art performance, significantly outperforming existing baselines. Our work not only provides a valuable resource for future research but also advances the field by offering a robust baseline for answering complex spatio-temporal questions.

\end{abstract}



\end{frontmatter}


\section{Introduction}
\label{intro}

Traditional knowledge graphs (KGs) represent static facts as triples (\emph{subject}, \emph{relation}, \emph{object}), lacking the ability to capture dynamic temporal changes or spatial relationships. In contrast, spatio-temporal knowledge graphs (STKGs), such as YAGO2~\cite{hoffart2013yago2}, enhance fact representation by incorporating timestamp and geographical coordinate annotations into their triples. This enables STKGs to capture dynamic spatio-temporal information, which is crucial for applications such as travel recommendation, historical event analysis, and urban planning~\cite{zhang2021framework,gottschalk2019eventkg,zhao2020urban}. Leveraging these advancements, researchers can now address questions that require both temporal and spatial reasoning, a task we define as spatio-temporal knowledge graph question answering (STKGQA).

\begin{figure}[htbp]
  \centering
    \includegraphics[width=0.9\linewidth]{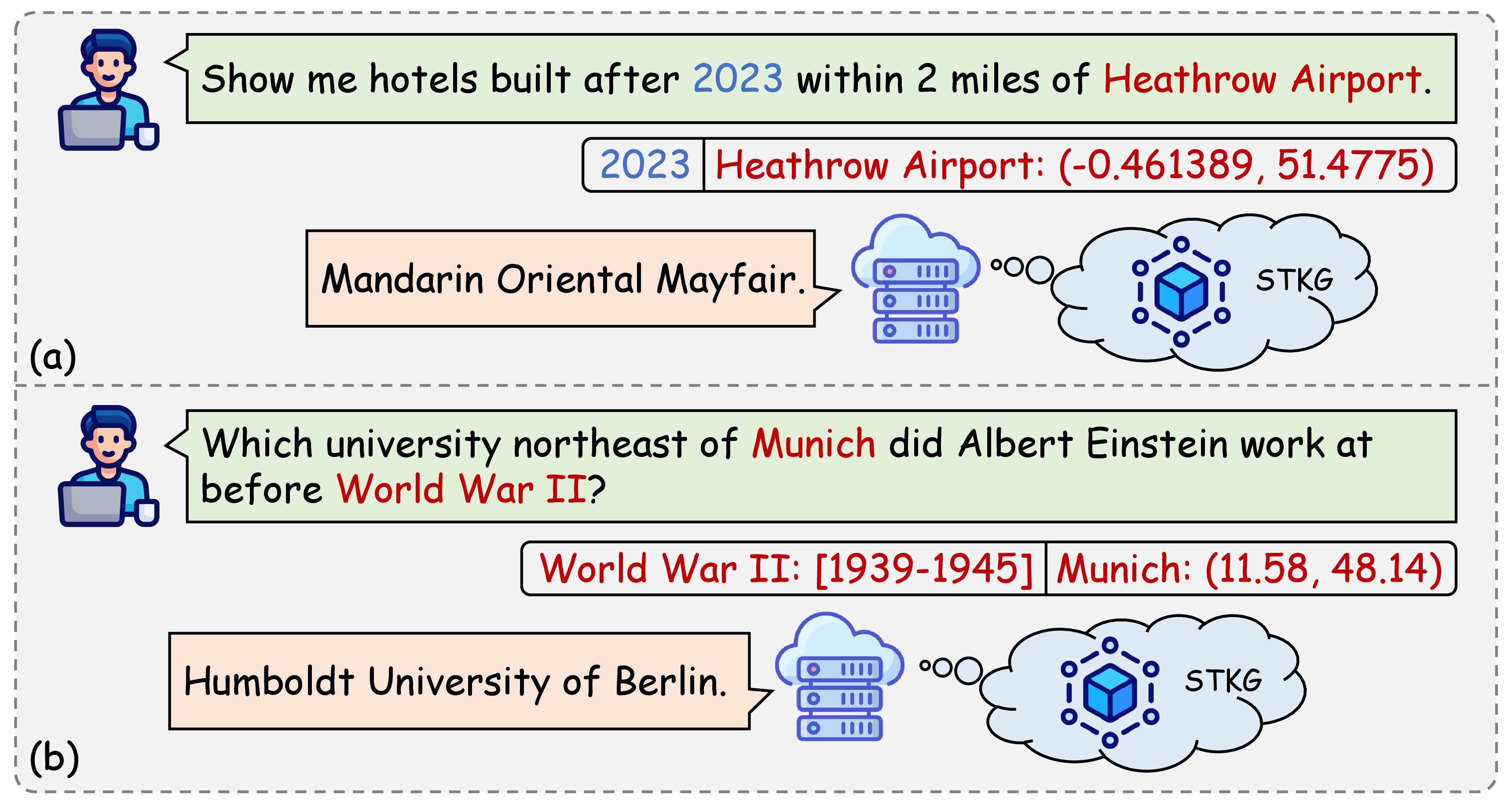}
\caption{The user request in Figure 1(a) involves the calculation of distance, while Figure 1(b) addresses the comparison of orientation. In the question presented in Figure 1(a), temporal information \emph{2023} is explicitly provided (in blue). The question in Figure 1(b) is more challenging, as neither the timestamp nor the geographic coordinates are explicitly provided (in red). 
}
\label{fig1}
\end{figure}

In fact, STKGQA is \textbf{widely demanded} and \textbf{highly challenging}. (1) As illustrated in Figure 1(a), a travel application service user may need to find recently opened hotels near an airport. The service provider needs to search its STKG for each hotel's establishment time and geographic coordinates and then calculate the distance to recommend hotels. Such complex queries are widely demanded in their business scenarios. (2) As shown in Figure 1(b), neither temporal or spatial information is explicitly given in the question. Correctly answering the question requires detecting implicit timestamps and geographic coordinates contained in entities (such as \emph{World War II} and \emph{Munich}) and reasoning on STKG. Therefore, it is a significant challenge to answer such complex questions accurately.

Despite its potential, current research on STKGQA is sparse and largely confined to a few case studies~\cite{hoffart2013yago2,stringhini2019spatio}.
As indicated in Table~\ref{tab1}, existing datasets either lack temporal annotations or spatial annotations, with no dataset supporting both spatial and temporal reasoning simultaneously. For instance, MULTITQ~\cite{chen2023multi}, one of the largest temporal KGQA datasets, contains 500k questions but lacks spatial annotations, while GeoQuestions1089~\cite{kefalidis2023benchmarking}, a spatial KGQA dataset, comprises 1k questions but ignores temporal reasoning. Hence, this data scarcity severely hinders the development of robust STKGQA methods.

\begin{table*}[t]
\centering
\resizebox{\columnwidth}{!}{%
\begin{tabular}{lccccc}
\toprule
Dataset & KG & Temporal & Spatial & Paraphrase & \#Question \\ \midrule
ComplexQuestions~\cite{bao2016constraint} & FreeBase & $\times$ & $\times$ & $\checkmark$ & 2k \\
TempQuestions~\cite{jia2018tempquestions} & FreeBase & $\checkmark$ & $\times$ & $\times$ & 1k \\
CronQuestions~\cite{saxena2021question} & Wikidata & $\checkmark$ & $\times$ & $\checkmark$ & 410k \\
MULTITQ~\cite{chen2023multi} & ICEWS & $\checkmark$ & $\times$ & $\times$ & 500k \\
GeoQuery~\cite{davis2007geoquery} & Geoquery Database & $\times$ & $\checkmark$ & $\times$ & 0.9k \\
GeoQuestions201~\cite{punjani2018template} & GADM, OSM, DBpedia & $\times$ & $\checkmark$ & $\times$ & 0.2k \\
GeoQuestions1089~\cite{kefalidis2024question} & YAGO2 & $\times$ & $\checkmark$ & $\checkmark$ & 1k \\ \midrule
STQAD (ours) & YAGO2 & $\checkmark$ & $\checkmark$ & $\checkmark$ & 10k \\ \bottomrule
\end{tabular}
}
\caption{KGQA dataset comparison. There is currently no dataset that contains both temporal and spatial information in one dataset.}
\label{tab1}
\end{table*}

To address this gap, we introduce STQAD, the first STKGQA dataset that includes an STKG with 15k entities and 138k facts, along with 10k natural language questions requiring spatio-temporal reasoning. STQAD is designed to reflect real-world scenarios, ensuring its practical relevance and utility for advancing STKGQA research. Specifically, we first align the temporal KG YAGO15K~\cite{GarcaDurn2018LearningSE} with the latest version of YAGO~\cite{suchanek2024yago} that includes timestamps and geographic coordinates, thus creating an STKG.
Based on pre-defined seed templates, we first select central entities containing spatio-temporal information from the STKG. Using these entities as the center, we expand the 2-hop range subgraphs to extract facts, ultimately generating template questions that contain implicit spatio-temporal information.
Our selection methods provide more relevant facts, ensuring our questions are more consistent with real-world use cases. To enhance the template questions' fluency and diversity, we use ChatGPT\footnote{https://openai.com/index/chatgpt} for paraphrasing, followed by rigorous validation to ensure the quality of all QA pairs.

On STQAD, we apply various language models~\cite{devlin2018bert,liu2019roberta,sanh2019distilbert,dubey2024llama} and KGQA methods~\cite{hu2017answering,saxena2021question,mavromatis2022tempoqr,chen2022temporal,chen2023multi} for evaluating their performance on the STKGQA task.
However, these methods struggle with the STKGQA task due to their lack of understanding and representation of spatio-temporal information.
To provide a competitive baseline for STQAD, we propose a spatio-temporal complex question answering (STCQA) method to address these questions effectively. 
STCQA consists of three key modules: preprocessing, answer retrieval, and answer filtering. In the preprocessing module, we identify spatio-temporal information in the question and retrieve their corresponding embeddings from the STKG. In the answer retrieval module, we extend the temporal KG embedding method to incorporate spatial information, enabling the joint representation of temporal and spatial features. The question is encoded using BERT~\cite{devlin2018bert}, and its embeddings are fused with the spatio-temporal embeddings from the STKG. A scoring function is then applied to retrieve preliminary answers. Finally, in the answer filtering module, we dynamically filter the retrieved answers based on spatio-temporal constraints, such as temporal comparisons (e.g., \emph{before}) and spatial distance calculations (e.g., \emph{within 3 miles}). This approach enables STCQA to outperform existing baselines by effectively modelling spatio-temporal interactions and handling constraints\footnote{Our code is available at https://github.com/OBriennnnn/STKGQA}. In summary, our contributions are as follows:

\begin{enumerate}
\item We introduce the STQAD dataset, consisting of 10k questions designed for the STKGQA task. To our knowledge, this is the first comprehensive KGQA dataset containing spatial and temporal information in each question.
\item We propose the STCQA to address STKGQA. This method can capture spatio-temporal information in the question and employ reasoning on STKG.
\item We demonstrate the quality of our new dataset and the effectiveness of our QA methods through comprehensive experiments. Furthermore, although we have obtained promising initial results, our dataset still offers ample opportunities for enhancing STKGQA.
\end{enumerate}

\section{Related Works}
\label{related_works}

\subsection{KGQA Datasets}
\label{qa_datasets}

Datasets are crucial for enhancing the generalization capabilities of KGQA models. However, as shown in Table~\ref{tab1}, existing datasets predominantly focus on either temporal or spatial questions without considering the more complex scenario of joint spatio-temporal reasoning.

To address the need for temporal reasoning, several temporal QA datasets have been developed. TORQUE~\cite{ning2020torque} introduces multiple-choice temporal questions with context, but its focus on multiple-choice format limits its applicability to open-ended QA tasks. TempQuestions~\cite{jia2018tempquestions} is a dataset specifically designed for temporal questions, containing 1,271 QA pairs collected through keyword filtering from WebQuestions~\cite{berant2013semantic}, Free917~\cite{cai2013large}, and ComplexQuestions~\cite{bao2016constraint}. However, its small size restricts its utility for spatio-temporal reasoning. CronQuestions~\cite{saxena2021question} is a large-scale temporal KGQA dataset comprising 410k questions, all requiring temporal KG to reason answers. While it significantly advances temporal QA research, it ignores spatial reasoning, limiting its applicability to real-world scenarios. MULTITQ~\cite{chen2023multi} further refines temporal granularity by incorporating questions with multiple timestamp levels, yet it also lacks spatial annotations. 

Furthermore, some QA datasets only focus on spatial reasoning. GeoQuery~\cite{davis2007geoquery} contains 880 hand-crafted factual questions about the geography of the United States. GeoAnQu~\cite{xu2020extracting} is a corpus of 429 complex and non-factual questions for geographic analysis, which are manually extracted from research papers containing GIS analysis and GIScience textbooks. GeoQuestions201~\cite{punjani2018template} is created for the evaluation of GeoQA, targeting linked geospatial datasets built from DBpedia and GADM and OSM datasets (UK and Ireland only), containing 201 manually crafted factual, simple, and complex questions. GeoQuestions1089~\cite{kefalidis2024question} is the largest KGQA benchmark that focuses on the geospatial domain of a specific KG and contains factual, simple, and complex questions that contain aggregates and superordinate levels. Despite these advancements, these datasets currently lack temporal annotations, underscoring the need for STQAD.

\subsection{KGQA Methods}
\label{kgqa_methods}

Recent approaches to KGQA are primarily designed for traditional KGs, where facts are represented as static triples. Some methods employ query graph extraction techniques to address KGQA tasks~\cite{yih2015semantic,bao2016constraint}, while others utilize neural network models to rank candidate answers or KG queries by learning scoring functions~\cite{hao2017end,lukovnikov2017neural,fevry2020entities,ye-etal-2022-rng,huang2023question}. Embedding-based methods have also proven effective in tackling these tasks~\cite{hao2017end,lukovnikov2017neural,fevry2020entities}. However, these methods are not designed to handle spatio-temporal information, limiting their applicability to STKGQA. 

For temporal KGQA tasks, recent research employs KG embedding techniques to learn embeddings for entities, relations, and timestamps. Methods such as CronKGQA~\cite{saxena2021question} and TempoQR~\cite{mavromatis2022tempoqr} determine answers by evaluating the distance between time information embeddings and question embeddings. While these approaches advance temporal reasoning, they neglect spatial information, making them unsuitable for spatio-temporal questions. 

For spatial KGQA tasks related to geographic coordinates, current research designs various query languages executable on KGs. Methods such as GeoQA~\cite{punjani2018template}, GeoQA2~\cite{punjani2023question}, and EarthQA~\cite{punjani2023earthqa} first extract entity and relation information from the questions, then fill them into the slots in query templates that can be executed on KGs. These approaches rely on pre-defined rules to map natural language questions to handcrafted SPARQL query templates, which limits their flexibility.

In STKGQA tasks, some studies~\cite{hoffart2013yago2,yin2019nlp} briefly discuss leveraging spatio-temporal information for reasoning on STKGs. However, these studies primarily rely on expert-crafted templates, which lack flexibility and scalability. Our method, STCQA, overcomes these limitations by dynamically processing spatio-temporal information and utilizing language models to comprehend question semantics, enabling robust reasoning over both temporal and spatial constraints.

\section{Preliminaries}
\label{preliminaries}

\subsection{STKG Definition}
\label{stkg_definition}

An STKG $\mathcal{K}:=(\mathcal{E}, \mathcal{R}, \mathcal{T}, \mathcal{L}, \mathcal{F})$ contains a set of entities $\mathcal{E}$, a set of relations $\mathcal{R}$, a set of timestamps $\mathcal{T}$, a set of locations $\mathcal{L}$, and a set of facts $\mathcal{F}$~\cite{hoffart2013yago2}. Each fact $(s, r, o, t, l) \in \mathcal{F}$ is a tuple where $s, o \in \mathcal{E}$ denote the subject and object entities, respectively, $r \in \mathcal{R}$ denotes the relation between them, $t \in \mathcal{T}$ is the timestamp of the fact and $l \in \mathcal{L}$ is the geographic coordinates of the object entity $o$. The timestamp $t$ of each fact is divided into a start time and an end time. Hence, a fact in STKG is represented as (\emph{subject, relation, object, occursSince, start\_time, occursUntil, end\_time, occursIn, <longitude, latitude>}), where \emph{occursSince}, \emph{occursUntil} and \emph{occursIn} are the start time, end time and location identifiers.

\subsection{Spatio-temporal Knowledge Graph Question Answering}
\label{spatio_temporal_question}

In STKGQA, answering spatio-temporal questions requires reasoning on STKG based on temporal or spatial information. We define spatio-temporal information in the question as comprising 2 primary components: \textbf{spatio-temporal clues} and \textbf{spatio-temporal constraints}. Spatio-temporal clues are entities or facts that contain timestamps or geographic coordinates, such as \emph{World War II} or \emph{Munich}. Spatio-temporal constraints are the reasoning rules based on timestamps and geographic coordinates, such as \emph{before}, \emph{northeast}, or \emph{within 3 miles}. Formally, the STKGQA task can be defined as follows: given a natural language question $\mathcal{Q}$ and an STKG $\mathcal{K}$, the goal is to retrieve a set of entities $\mathcal{A} \subseteq \mathcal{E}$ that satisfy the spatio-temporal constraints embedded in $\mathcal{Q}$.

\section{The STQAD Dataset}
\label{the_stqad_dataset}

The construction of STQAD follows 4 main \textbf{Guidelines}:
(1) The questions must be based on STKG and involve temporal and spatial reasoning.
(2) Spatio-temporal information should not be explicitly stated in the questions, as users typically cannot provide precise timestamps or geographic coordinates.
(3) The questions should resemble those posed by humans and reflect real-world scenarios.
For example, consider the following questions: \emph{Which university northeast of Munich did Einstein work at before \textbf{World War II?}} and \emph{Which university northeast of Munich did Einstein work at before \textbf{the 25th Academy Awards}?} Clearly, the temporal information in the former question is more related to \emph{Einstein}.
(4) The size of the dataset must be sufficient to train the model.

STQAD comprises 2 main parts: an STKG with spatio-temporal annotations and a set of QA pairs that require spatio-temporal reasoning. We enhance the original temporal KG~\cite{GarcaDurn2018LearningSE} by adding timestamps and geographic coordinates, expanding it into an STKG. Based on STKG, we develop 40 seed templates and then fill out the spatio-temporal clues and constraints to generate 10k template questions. In order to improve the diversity and fluency of the template questions, we use ChatGPT to paraphrase them. Finally, all answers, clues, and constraints are re-verified to ensure the quality of the QA pairs.

\subsection{STKG Construction}
\label{stkg_construction}

YAGO15k~\cite{GarcaDurn2018LearningSE} is a small-scale temporal KG extracted from the standard YAGO, with facts annotated with timestamps. Unlike YAGO15k, YAGO2~\cite{hoffart2013yago2} is specifically designed to represent spatio-temporal information for facts. Given that YAGO15k already contains temporal annotations, we believe it provides a suitable foundation for integrating spatial information from YAGO2, thereby extending it into an STKG. To enhance the spatial information in YAGO15k, we supplement its facts with spatio-temporal annotations from YAGO2. Specifically, we extract geographic coordinates and timestamps from YAGO2 and align them with the corresponding entities and facts in YAGO15k.

\begin{table}[htbp]
\centering
\renewcommand{\arraystretch}{0.85}
\resizebox{0.6\columnwidth}{!}{%
\begin{tabular}{lc}
    \toprule
    Statistics type&Quantity\\
    \midrule
    Facts&138,056\\
    Facts (spatio-temporal annotation)&42,185\\
    Entities&15,403\\
    Relations&113\\
    Timestamps&572\\
    Geographic coordinates&2,235\\
    Time span (year)&2-3150\\
    \bottomrule
\end{tabular}
}
\caption{Statistics for our STKG.}
\label{tab2}
\end{table}

For facts that already contain timestamps or geographic coordinates in YAGO, we directly assign these annotations to the corresponding facts in YAGO15k. However, some facts in YAGO15k lack geographical information. In such cases, we supplement the missing location data by using the geographic coordinates of the \emph{object} entity in the fact. For example, considering a fact (\emph{Albert\_Einstein, worksAt, Humboldt\_University\_of\_Berlin, occursSince, 1914, occursUntil, 1917}), we can assign the location coordinates of \emph{Humboldt\_University\_of\_Berlin} as \emph{\textless 52.52, 13.39 \textgreater} to complete the fact. This process ensures that many facts in the STKG are enriched with both temporal and spatial information. All the facts in the STKG are then used as the basis for generating questions, and Table~\ref{tab2} shows the statistics.

\subsection{QA Dataset Generation}
\label{qa_dataset_generation}

In order to construct QA pairs, we first design 40 seed templates with executable procedures based on the 10 most frequent STKG relations in YAGO15K (Section~\ref{seed_template_construction}).
Next, we select facts with spatio-temporal information from STKG and expand them into the 2-hop subgraphs for extracting spatio-temporal clues.
By comparing these facts and clues, we get the spatio-temporal constraints (Section~\ref{spatio_temporal_clue_and_constraint_generation}). Finally, we fill the seed templates with these clues and constraints, enriching and diversifying the templates' semantics through paraphrasing with ChatGPT. Additionally, we validate the effectiveness of the spatio-temporal constraints and generate the final answers(Section~\ref{qa_pair_generation}). This approach constructs 10k QA pairs, with each question's entities annotated and linked to our STKG.

\subsubsection{Seed Template Construction}
\label{seed_template_construction}

We select the 10 most common KG relations involving both time and location as types of questions. These relations are as follows: \emph{location}, \emph{livesIn}, \emph{memberOf}, \emph{dealsWith}, \emph{playsFor}, \emph{worksAt}, \emph{isCitizenOf}, \emph{hasNeighbour}, \emph{isPoliticianOf}, \emph{graduatedFrom}. Each relation corresponds to 4 seed templates that include spatio-temporal reasoning (please see some examples in Table~\ref{tab3}). 
Each template is associated with a corresponding seed procedure, which is essentially a SPARQL template containing placeholders for variables and spatio-temporal information. 
Once we determine spatio-temporal clues and constraints for these templates, the procedure can be executed and search for all possible answers from the STKG (\ref{seed_procedure} shows more details). 
\begin{table}[htbp]
\resizebox{\columnwidth}{!}{%
\small
\begin{tabular}{>{\centering\arraybackslash}p{2.5cm}p{6cm}p{6.3cm}}
\toprule
Relation & \centering\arraybackslash{Seed Template} & \centering\arraybackslash{Template Question} \\ 
  
\cmidrule(r){1-1} \cmidrule(r){2-2} \cmidrule(r){3-3}
  
\textless{}playsFor\textgreater{} &
  {[}\textbf{\emph{geo constraint}}{]} \{\textbf{\emph{geo clue}}\}, which team did \{\textbf{\emph{central entity}}\} play for {[}\textbf{\emph{temporal constraint}}{]} \{\textbf{\emph{temporal clue}}\}? &
  \{Southwest of\} \textless{}Stockholm\textgreater{}, which team did \textless{}George\_Moncur\textgreater play for \{after\} \textless{}Stockport\_County\_F.C.\textgreater{}? \\
  
\cmidrule(r){1-1} \cmidrule(r){2-2} \cmidrule(r){3-3}
  
\textless{}hasNeighbor\textgreater{} &
  Name all countries created {[}\textbf{\emph{temporal constraint}}{]} \{\textbf{\emph{temporal clue}}\} and {[}\textbf{\emph{geo constraint}}{]} \{\textbf{\emph{geo clue\&central entity}}\}. &
  Name all countries created \{before\} \textless{}Trinity\_College\_(Connecticut)\textgreater and \{southeast of\} \textless{}Algeria\textgreater{}. \\

\cmidrule(r){1-1} \cmidrule(r){2-2} \cmidrule(r){3-3}
  
\textless{}location\textgreater{} &
  What hotels in \{\textbf{\emph{central entity}}\} were built {[}\textbf{\emph{temporal constraint}}{]} \{\textbf{\emph{temporal clue}}\} {[}\textbf{\emph{geo constraint}}{]} \{\textbf{\emph{geo clue}}\}? &
  What hotels in \textless{}London\textgreater{} were built \{after\} \textless{}London\_Olympic\_Games \textgreater{} \{within 3 miles of\} \textless{}Big\_Ben\textgreater{}? \\ 
  \bottomrule
\end{tabular}%
}
\caption{Some representative template questions generated from seed templates. In some seed templates, the \textbf{\emph{central entity}} overlaps with the \textbf{\emph{geo clues} (\{\textbf{\emph{geo clue\&central entity}}\}}). All relations have seed templates for orientation comparison. In addition, the seed templates corresponding to some relations (such as \emph{location}) also involve distance calculation on the geographic constraint.
\ref{more_statistics_and_details_of_stqad} shows more seed template examples. 
}
\label{tab3}
\end{table}

As shown in Table~\ref{tab3}, the seed templates involve three types of information that need to be filled in: central entity, spatio-temporal clues, and spatio-temporal constraints. (1) \textbf{Central entity} refers to the subject entity \( s \) or object entity \( o \) in the STKG fact. (2) \textbf{Spatio-temporal clues} are presented as events or entities associated with the central entity and contain implicit spatio-temporal information. (3)  \textbf{Spatio-temporal constraints} guide the reasoning about the central entity based on the spatio-temporal clues.

\subsubsection{Spatio-temporal Clue and Constraint Generation}
\label{spatio_temporal_clue_and_constraint_generation}

\begin{figure}[t]
  \centering
  \includegraphics[width=\textwidth]{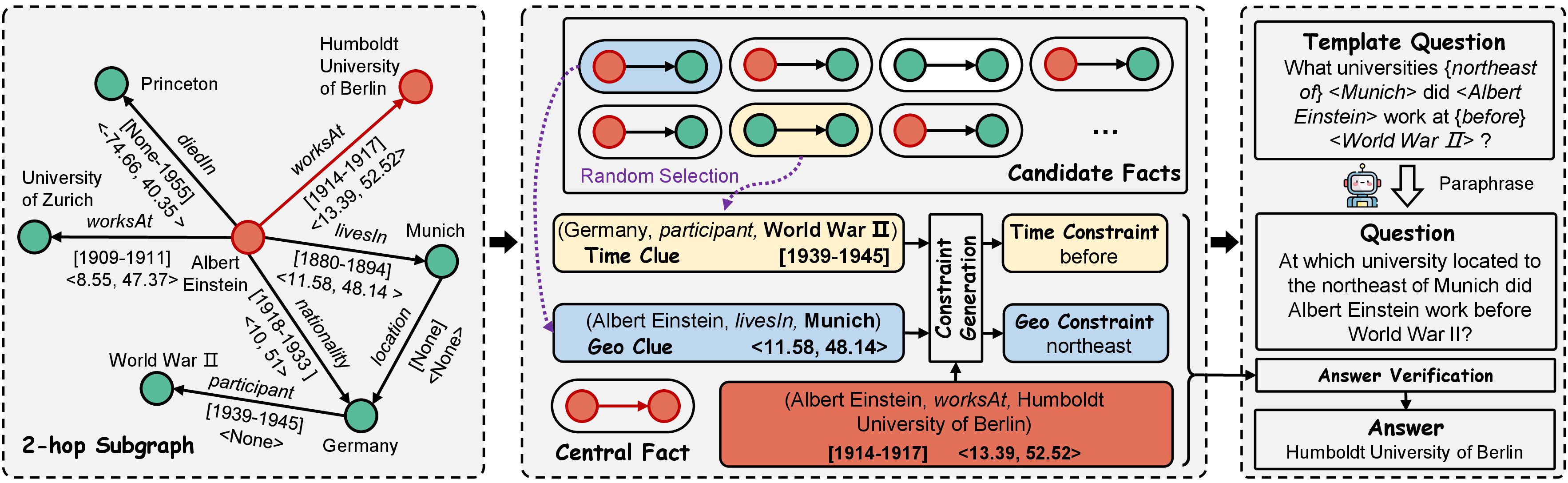}
  \caption{The generation of QA pairs. If the fact in the 2-hop subgraph lacks spatio-temporal information, we label it as \emph{None}. We randomly select facts containing spatial or temporal information to serve as clues and generate constraints by comparing them with the \emph{Central Fact} (in red). During the \emph{Answer Verification} process, we examine the generated questions and answers by executable procedure to confirm the validity of the clues and constraints, ensuring the quality of the QA pairs.}
  \label{fig2}
\end{figure}

Figure~\ref{fig2} illustrates the process of generating spatio-temporal clues and constraints. For the generation of clues, we use the fact (\emph{Albert Einstein, worksAt, Humboldt University of Berlin, occursSince, 1914, occursUntil, 1917, occursIn, <52.52, 13.39>}) as central fact, where \emph{Albert Einstein} is the central entity and \emph{Humboldt University of Berlin} is the candidate answer. To facilitate the generation of spatio-temporal clues, we search STKG for all facts related to \emph{Albert Einstein} within 2 hops, which contain temporal or geographical information.
These facts neighboring the central entity are regarded as highly relevant clue candidates. An example of such a fact is (\emph{Albert Einstein, livesIn, Munich, occursSince, 1880, occursUntil, 1894, occursIn, <11.58, 48.14>}). Furthermore, some ambiguous relations (such as \emph{influence} and \emph{linksTo}) in these facts may affect the semantic quality of the template question, so we filter them out.
From the remaining facts, we randomly select one fact containing a timestamp as the \emph{Time Clue}, and then randomly choose another fact containing geographic coordinates as the \emph{Geo Clue}.

To generate spatio-temporal constraints, we compare the spatio-temporal information contained in the clues with that of the central fact. Temporal constraints are categorized into 2 types: Double Timestamp Constraints (DTC) and Single Timestamp Constraints (STC). 
DTC, such as \emph{during} and \emph{while}, require that the start and end timestamp of the central fact are entirely contained within the start and end intervals of the \emph{Time Clue}. DTC is determined through twice comparisons. 
STC, such as \emph{before}, requires that the end timestamp of the central fact precedes the start timestamp of the \emph{Time Clue}, while  \emph{after} requires that the start timestamp of the central fact follows the end timestamp of the\emph{Time Clue}. STC is determined through a single comparison. 
Since facts that satisfy DTC are uncommon in our STKG, we prioritize generating these constraints to ensure a balanced distribution of constraint types.

Spatial constraints include 3 types: Double Direction Constraint (DDC), Single Direction Constraint (SDC), and Distance Constraint (DC). DDC, such as \emph{northeast}, require simultaneous comparisons of both latitude and longitude. SDC, such as \emph{north} or \emph{east}, require a comparison of either latitude or longitude. For DC, we designed a geographic coordinates calculator to compute distances and ceiling the result. For example, if the calculation yields \emph{2.4 miles}, the spatial constraint is \emph{3 miles}. 
Since latitude and longitude annotations for facts are more prevalent in STKG, these 3 types of spatial constraints are selected randomly.

\begin{table}[htbp]
\centering
\resizebox{9cm}{!}{%
\begin{tabular}{lccc}
\toprule
                            & \textbf{Train} & \textbf{Dev} & \textbf{Test} \\ 
\cmidrule(r){1-1} \cmidrule(r){2-4}
Double Timestamp Constraint (DTC) & 281            & 36           & 28            \\ 
Single Timestamp Constraint (STC) & 8,224           & 1,027          & 1,035           \\
\cmidrule(r){1-1} \cmidrule(r){2-4}                                      
Double Direction Constraint (DDC) & 2,861           & 363          & 346           \\ 
Single Direction Constraint (SDC) & 2,852           & 356          & 362           \\
Distance Constraint (DC) & 2,792           & 344          & 355           \\ 
\cmidrule(r){1-1} \cmidrule(r){2-4}
Overall Number              & 8,505           & 1,063         & 1,063          \\
Average Sentence Length     & 16.64          & 16.68        & 16.63         \\
\bottomrule
\end{tabular}%
}
\caption{Statistics of the STQAD. More statistics are listed in~\ref{more_statistics_and_details_of_stqad}.}
\label{tab4}
\end{table}

\subsubsection{QA Pair Generation}
\label{qa_pair_generation}

After obtaining all necessary elements for question generation, we fill the central entities, spatio-temporal clues, and constraints into seed templates to create template questions.
These questions are then paraphrased using ChatGPT to enhance their diversity and fluency, and they are manually verified for correctness.
Furthermore, verifying the answers is crucial to ensuring the quality of the QA pairs: (1) the answer from the central fact may only be one of the correct answers; (2) it is essential to determine whether the spatio-temporal constraints are effective, specifically whether these constraints help narrow the search space when querying answers in the STKG.

We first re-retrieve the answers using the executable procedure from the seed template to obtain all correct answers. This step is necessary because the spatio-temporal constraints are derived from the central fact in a backward manner, ensuring that the \emph{object} of the central fact is included in the answer set. However, it may not be the only correct answer. Therefore, we further utilize the procedure to retrieve answers from the STKG that satisfy all the constraints.
Next, we remove the spatio-temporal clues and constraints from the procedure and observe whether the number of retrieved answers increases. 
If removing the constraints does not increase the number of retrieved answers, we consider the constraints ineffective for narrowing the search space in the STKG; such QA pairs are therefore discarded from the final dataset.
With this approach, we generate 10k QA pairs and then split them into training, validation, and test sets in a ratio of 8:1:1.

\begin{figure}[htbp]
  \centering
    \includegraphics[width=0.84\textwidth]{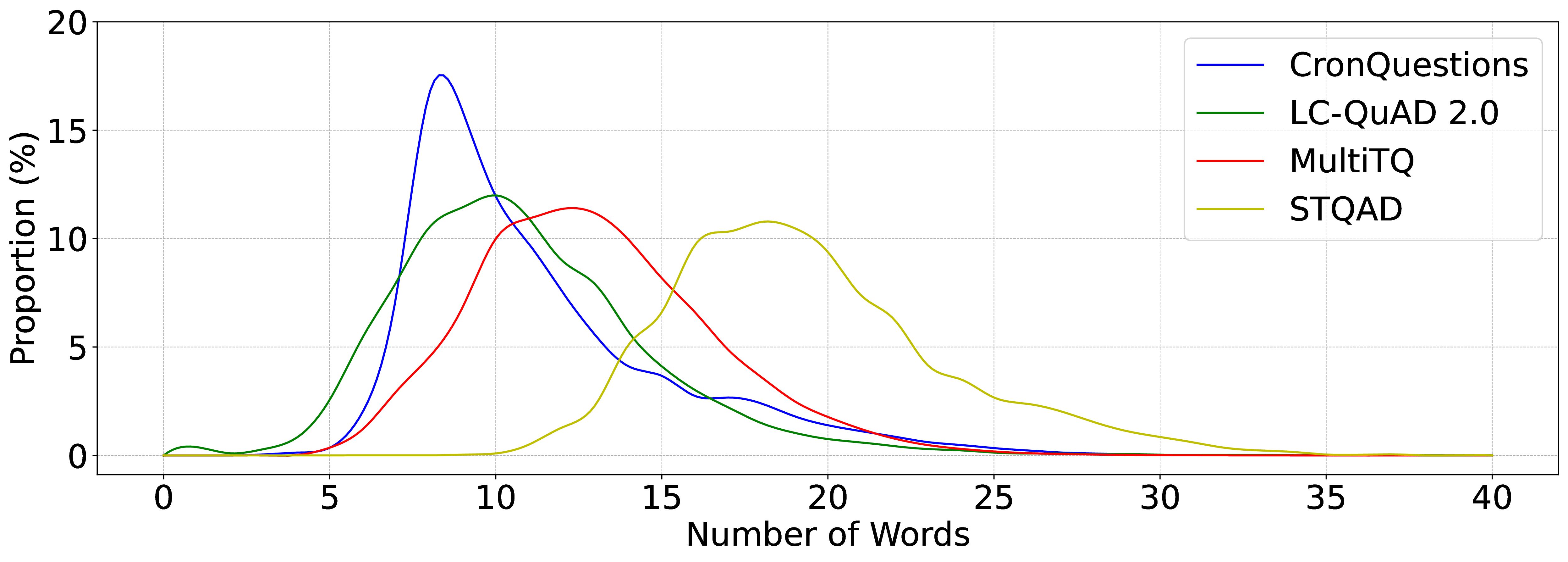}
  \caption{Length distribution of datasets.}
\label{fig3}
\end{figure}

Table~\ref{tab4} provides the statistics of our dataset. Each question in STQAD involves multiple entities and relations, and neither timestamps nor geographic coordinates are explicitly provided. In Figure~\ref{fig3}, we present the distribution of question lengths (measured in word count) from STQAD and compare it with several current datasets. The figure indicates that a good proportion of questions in STQAD are relatively verbose, which increases the complexity for QA systems to parse.

\section{STCQA method}
\label{stcqa_method}

\subsection{Overview}
\label{overview}

\begin{figure}[t]
  \centering
  \includegraphics[width=\textwidth]{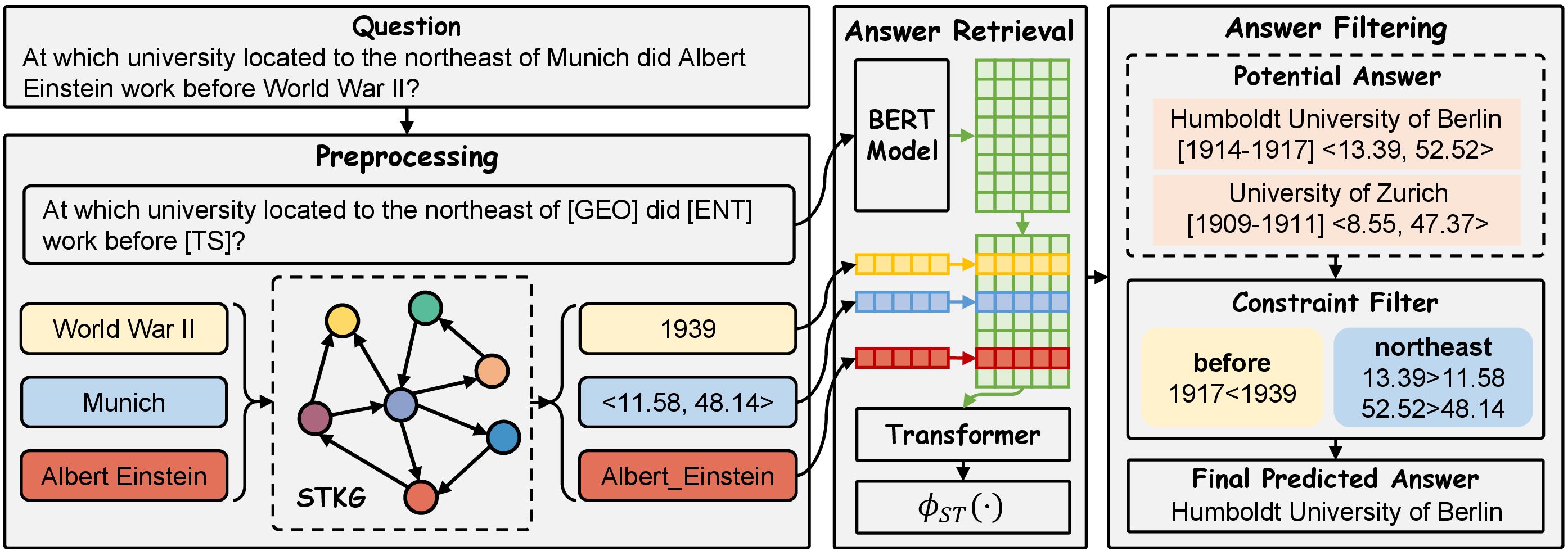}
  \caption{Model architecture of STCQA.}
  \label{fig4}
\end{figure}

The overall architecture of STCQA, as shown in Figure \ref{fig4}, consists of 3 modules: Preprocessing, Answer Retrieval, and Answer Filtering.
In the Preprocessing module, we annotate the attributes of facts or entities in the question to identify the central entity and spatio-temporal clues. Then, we retrieve corresponding timestamps and geographic coordinates of spatio-temporal clues from the STKG. The central entity and spatio-temporal clues are replaced with special tokens to minimize their impact on the question's logical semantics.
In the Answer Retrieval module, we replace the special tokens' embeddings in the BERT-encoded~\cite{devlin2018bert} question with the spatio-temporal clues' embeddings from the STKG. We then use a 2-layer transformer to fuse question semantic encoding and apply the scoring function to retrieve all potential answers from the STKG.
In the Answer Filtering module, we apply spatio-temporal constraints to eliminate incorrect potential answers, thereby obtaining the final answers.

\subsection{Preprocessing Module}
\label{preprocessing_module}

To extract entities from the spatio-temporal questions, we employ the named entity recognition tool FLERT~\cite{schweter2020flert} and the entity linking tool BLINK~\cite{wu2019scalable}. These tools enable the alignment of entities mentioned in the questions with those in the STKG.
The entities involved in a question are categorized into three types: central entities, entities implying temporal information, and entities implying geographical information. The determination of entity types relies on constraint keywords that precede the entities, such as \emph{before}, \emph{northeast of}, and \emph{within 3 miles of}.
We use special tokens \texttt{[ENT]}, \texttt{[TS]} and \texttt{[GEO]} to mark entities: \texttt{[ENT]} denotes the central entity, \texttt{[TS]} denotes an entity that primarily provides a temporal clue, and \texttt{[GEO]} denotes an entity that primarily provides a spatial clue.
Subsequently, based on the classified special tokens, we retrieve precise timestamps or geographic coordinates contained in these entities from the STKG.

\subsection{Answer Retrieval Module}
\label{answer_retrieval_module}

Firstly, we replace the special token embeddings with their respective STKG embeddings. To encode entities, timestamps and geographic coordinates in STKG, we extend the geographical information in embedding method TComplEx~\cite{lacroix2020tensor}. TComplEx, a temporal extension of ComplEx~\cite{trouillon2016complex}, represents each timestamp \(t\) as a complex vector \(\mathbf{t}\), and the score \({\phi}_T\) for a fact \((s, r, o, t)\) is defined as

\begin{equation}
{\phi}_T\left(\mathbf{e}_s, \mathbf{r}, \overline{\mathbf{e}}_o, \mathbf{t}\right)=\operatorname{Re}\left(\left\langle\mathbf{e}_s, \mathbf{r} \odot \mathbf{t}, \overline{\mathbf{e}}_o\right\rangle\right)
\end{equation}
where $\mathbf{e}_s$, $\mathbf{r}$, $\mathbf{t}$ are the embedding representations of \(s, r, t\). $\overline{\mathbf{e}}_o$ is the complex vector of $o$, and $\odot$ is the element-wise product. 
We extend TComplEx to incorporate location information. The scoring function is defined as 
\begin{equation}
{\phi}_{ST}\left(\mathbf{e}_s, \mathbf{r}, \overline{\mathbf{e}}_o, \mathbf{t}\right)=\operatorname{Re}\left(\left\langle\mathbf{e}_s, \mathbf{r} \odot \mathbf{t} \odot \mathbf{l}, \overline{\mathbf{e}}_o\right\rangle\right)
\end{equation}
where the location $l$ is represented as the vector $\mathbf{l}$. All embedding vectors are learned such that the scoring function ${\phi}_{ST}$ assigns higher scores to valid facts $(s, r, o, t, l) \in \mathcal{F}$ compared to invalid facts $\left(s^{\prime}, r^{\prime}, o^{\prime}, t^{\prime}, l^{\prime}\right) \notin \mathcal{F}$. 
Formally, we require
\begin{equation}
{\phi}_{ST}\left(\mathbf{e}_s, \mathbf{r}, \overline{\mathbf{e}}_o, \mathbf{t}, \mathbf{l}\right) > {\phi}_{ST}\left(\mathbf{e}_s^{\prime}, \mathbf{r}^{\prime}, \overline{\mathbf{e}}_o^{\prime}, \mathbf{t}^{\prime}, \mathbf{l}^{\prime} \right)
\end{equation}
where $\mathbf{e}_s$ and $\mathbf{e}_s^{\prime}$ are subject embeddings, $\mathbf{r}$ and $\mathbf{r}^{\prime}$ are relation embeddings, $\overline{\mathbf{e}}_o$ and $\overline{\mathbf{e}}_o^{\prime}$ are object embeddings, and $\mathbf{t}$, $\mathbf{t}^{\prime}$, $\mathbf{l}$, $\mathbf{l}^{\prime}$ denote timestamps and geographic coordinates for valid and invalid facts, respectively.

To retrieve answers from the STKG, we encode the entities, timestamps, and geographic coordinates obtained during the Preprocessing stage and replace them with special tokens' corresponding positions in the BERT-encoded question embeddings. Then, we use a 2-layer Transformer~\cite{vaswani2017attention} to integrate all information. The question representation vector, denoted as $\mathbf{q}$, integrates the semantics of the question and spatio-temporal clues, which can better assist the model in reasoning on STKG. 
The final answer score is the element-wise maximum of two directed scores:

\begin{equation}
	\begin{split}
	    max ( & {\phi}_{ST}\left(\mathbf{e}_c, \mathbf{W}_E \mathbf{q}, \mathbf{e}_\mathcal{E}, \mathbf{v}_t, \mathbf{v}_l\right) ,\; {\phi}_{ST}\left(\mathbf{e}_\mathcal{E}, \mathbf{W}_E \mathbf{q}, \mathbf{e}_c, \mathbf{v}_t, \mathbf{v}_l\right) )
	\end{split}
\end{equation}
where $\mathbf{e}_c$ represents the embedding vector of the central entity identified in the preprocessing module, $\mathbf{W}_E$ is a $D \times D$ learnable matrix specific for answer prediction with $D$ being the embedding dimension, $\mathbf{q}$ denotes the final representation vector of the question obtained from the 2-layer Transformer, $\mathbf{e}_\mathcal{E}$ represents the embedding vectors of all candidate answer entities in the STKG, $\mathbf{v}_t$ and $\mathbf{v}_l$ are the embedding vectors of the timestamp and geographic coordinate derived from the temporal and spatial clues in the question respectively.
The scores for all entities, times and locations are concatenated, and the $Softmax$ function is used to calculate answer probabilities over this combined score vector. During training, the model updates its parameters by minimizing the cross-entropy loss function, aiming to assign a higher probability to the correct answer.

\subsection{Answer Filtering Module}
\label{answer_filter_module}

Although our scoring functions have retrieved some potential answers in the STKG, embedding-based KGQA methods exhibit limited spatio-temporal reasoning or distance calculation capabilities on STKG. To address this issue, we design a Constraint Filter within the Answer Filtering Module. This filter performs temporal and spatial comparisons or distance calculations based on the spatio-temporal constraints in the questions.
We implement explicit rules for temporal and orientation comparisons, and use the Haversine formula~\cite{robusto1957cosine} to compute geographic distances.

Specifically, we apply the constraints obtained from the Preprocessing Module to filter the potential answers and thus determine the final results. 
When a \textbf{directional constraint} appears in the question, the geographic coordinates of each candidate answer are compared with those of the spatial clue. In such comparisons, equality on target coordinate dimension does not satisfy the requirement; for example, the constraint \emph{northeast} demands that both the latitude and the longitude of the candidate's location exceed those of the spatial clue. 
In the case of a \textbf{distance constraint}, we compute the great-circle distance via the Haversine formula, round it to the nearest 0.1 mile, and consider the condition satisfied only when the result is \emph{less than or equal} to the specified threshold. 
For \textbf{temporal constraints}, a DTC such as \emph{during} is satisfied only when the candidate’s start and end times are entirely contained within the temporal clue’s time span. In STC, if the compared timestamps are equal (e.g., the candidate’s end time equals the clue’s start time), the condition is treated as unsatisfied. For example, as illustrated in Figure~\ref{fig4}, the \emph{before} constraint requires that the candidate’s end time occur strictly earlier than the clue’s start time.


\section{Experiments}
\label{experiments}

In this section, we conduct a comprehensive evaluation of the STQAD dataset and the STCQA method. Section ~\ref{experimental_setup} introduces the evaluation metrics and baselines. In Section~\ref{overall_results}, we evaluate the performance of the baselines and STCQA on the STQAD dataset, and we also observe \textbf{Guideline (1)} from Section~\ref{the_stqad_dataset} based on the overall results. Section~\ref{ablation_study} presents an ablation study to analyze the effectiveness of different modules within STCQA. Additionally, we observe \textbf{Guideline (2)} from Section~\ref{the_stqad_dataset} based on the ablation results. Section~\ref{stkg_embedding_performance} tests the performance of our embedding method on the STKG. In Section ~\ref{effect_of_spatio_temporal_constraint_relevance}, we assess the impact of factual relevance in our question on the QA system, thereby investigating \textbf{Guideline (3)} from Section~\ref{the_stqad_dataset}. Section~\ref{effect_of_dataset_size} examines how the scale of the STQAD dataset influences the performance of the QA system, testing \textbf{Guideline (4)}. Finally, Section~\ref{prompt_inject} evaluates the performance impact of LLM when injecting relevant KG facts into prompts.

\subsection{Experimental Setup}
\label{experimental_setup}

\subsubsection{Baselines}
\label{baselines}

\paragraph{Pre-trained language models (PLMs)} Research indicates that PLMs contain real-world knowledge and can be directly applied to QA tasks~\cite{petroni2019language,raffel2020exploring}. In our experiments, we evaluate the performance of three widely-used PLMs: BERT~\cite{devlin2018bert}, RoBERTa~\cite{liu2019roberta}, and DistillBERT~\cite{sanh2019distilbert} on the STQAD dataset. For each model, we use the pre-trained versions available in the Hugging Face Transformers library\footnote{https://huggingface.co}, with the base configuration (12 layers, 768 hidden dimensions). We concatenate the question embedding generated by the PLM with the spatio-temporal entity embeddings from the STKG. This concatenated representation is then passed through a learnable projection layer (a fully connected layer with 512 output dimensions) to align the dimensions of the question and entity embeddings. The resulting embedding is scored using the dot product over all entities in the STKG, and the top-k entities with the highest scores are selected as candidate answers.



\paragraph{Large language models (LLMs)} LLMs with billions of parameters demonstrate strong performance and achieve notable success in QA tasks. In our experiments, we ask ChatGPT, GPT-4o~\footnote{https://openai.com/index/hello-gpt-4o}, Llama3 70B~\cite{dubey2024llama}, GPT-4.5, Claude 3.7~\footnote{https://www.anthropic.com/news/claude-3-7-sonnet} and DeepSeek-V3~\cite{liu2024deepseek} to answer questions in STQAD. We input the questions directly into these models and request them to provide the 10 most probable answers in order of likelihood. If they cannot answer, we instruct them to respond with \emph{None}.
In evaluation, \cite{tan2023can} argues that for LLMs to achieve stable accuracy in KGQA tasks, a threshold of at least 0.7 should be applied when comparing generated outputs to ground truth. Following this guideline, we adopt a Levenshtein similarity threshold of 0.7 based on empirical tests to match candidate answers from YAGO2 with the aliases of the correct answers. To validate this setting, we sample 100 questions from the STQAD test set and manually compute matching accuracy under thresholds $\{0.6, 0.7, 0.8, 0.9\}$. We find that a threshold of $0.7$ yielded the best trade-off between recall and precision for accommodating entity alias variation in YAGO2 without over-matching unrelated entities. Thresholds below $0.7$ led to a sharp decline in precision due to excessive fuzzy matches.

\paragraph{KGQA models} EmbedKGQA~\cite{hu2017answering} is a framework for QA over traditional KG, so timestamps and coordinates are not used during the pre-training of the model.
CronKGQA~\cite{saxena2021question} employs a PLM to obtain question embeddings and utilizes a temporal KG embedding scoring function for answer prediction. In our experiment, we annotated all time-related entities and utilized the model to predict answers. TempoQR~\cite{mavromatis2022tempoqr} utilizes a temporal KG embedding-based scoring function for answer prediction and incorporates additional temporal information. Thus, we also incorporate time information in STKG into the central entity. SubGTR~\cite{chen2022temporal} extracts implicit information from temporal questions and incorporates a semantic-aware and temporal inference module into the scoring function, so we also provide time clues and time constraints for the model. MultiQA~\cite{chen2023multi} refines the granularity of time representation from \emph{Year} to \emph{Day}. Our experiment employed its embedding results for \emph{Year}. 
GeoQA2~\cite{punjani2023question} is also designed based on YAGO2 and can execute SPARQL queries on the terminal. We provide the spatio-temporal questions to the system, which then generates queries to find answers within YAGO2.
We provide entity annotations for all PLM and KGQA models, allowing the evaluation results to be independent of the performance of named entity recognition and entity linking tools. 
\ref{implementation_details} provides more details of experimental settings.

\subsubsection{Evaluation Metrics}
The KGQA approach typically follows a retrieval and ranking paradigm~\cite{miller2016key,zhang2018variational,saxena2020improving}. It constructs a subgraph from the KG and ranks all entities in the subgraph based on their relevance to the question. For fairness and standard of comparison with other QA methods, we adopted the hit ratios (HR) used in temporal KGQA task~\cite{saxena2021question,mavromatis2022tempoqr,chen2022temporal,chen2023multi} as the evaluation metric in our study. 

\subsection{Overall Results} 
\label{overall_results}

Table~\ref{tab5} presents a comparative analysis of STCQA against other baselines. Initially, by comparing PLMs with KGQA models utilizing embedding approaches (EmbedKGQA, TempoQR, SubGTR, and MultiQA), we observe that even without explicit spatio-temporal information, using KG embeddings significantly enhances model performance. For example, EmbedKGQA achieves a Hits@1 score of 37.16\%, compared to 30.20\% for BERT. This suggests that KG embeddings help models focus more effectively on sparsely distributed entities within the STKG, enabling better reasoning over the graph structure~\cite{hu2017answering}.

\begin{table}[t]
\centering
\renewcommand{\arraystretch}{1.1}
\resizebox{\columnwidth}{!}{%
\setlength{\tabcolsep}{2pt}
\begin{tabular}{llcccccccccccc}
\toprule
\multicolumn{2}{c}{\multirow{2}{*}{Models}} & \multicolumn{6}{c}{Hits@1} & \multicolumn{6}{c}{Hits@10} \\

\cmidrule(r){3-8} \cmidrule(r){9-14}

\multicolumn{2}{c}{} & \textbf{Overall} & DTC & STC & DDC & SDC & DC & \textbf{Overall} & DTC & STC & DDC & SDC & DC \\ 

\cmidrule(r){1-2} \cmidrule(r){3-3} \cmidrule(r){4-5} \cmidrule(r){6-8} \cmidrule(r){9-9} \cmidrule(r){10-11} \cmidrule(r){12-14}

\multirow{3}{*}{\emph{PLMs}} & BERT & 30.20 & 28.57 & 30.24 & 31.21 & 29.56 & 29.86 & 60.96 & 53.57 & 61.16 & 59.25 & 61.620& 61.97 \\
 & RoBERTa & 31.89 & 32.14 & 31.88 & 33.53 & 32.04 & 27.61 & 60.77 & 57.14 & 60.87 & 61.85 & 58.29 & 62.25 \\
 & DistillBERT & 29.63 & 25.00 & 29.76 & 30.06 & 27.62 & 29.30 & 58.23 & 53.57 & 58.36 & 57.23 & 57.18 & 60.28 \\

\cmidrule(r){1-2} \cmidrule(r){3-3} \cmidrule(r){4-5} \cmidrule(r){6-8} \cmidrule(r){9-9} \cmidrule(r){10-11} \cmidrule(r){12-14}
 
\multirow{6}{*}{\emph{LLMs}} & ChatGPT & 10.35 & 10.71 & 10.34 & 9.54 & 11.88 & 8.73 & 24.65 & 25.00 & 24.64 & 22.25 & 25.41 & 26.20 \\
 & GPT-4o & 13.73 & 14.29 & 13.72 & 15.32 & 14.36 & 10.42 & 25.21 & 28.57 & 25.12 & 27.75 & 25.97 & 24.97 \\
 & Llama3 & 9.13 & 7.14 & 9.18 & 9.83 & 9.39 & 7.60 & 15.43 & 17.86 & 15.36 & 16.47 & 15.47 & 14.37 \\ 
 & GPT-4.5 & 22.48 & 17.86 & 22.61 & 26.01 & 21.27 & 20.28 & 63.02 & 50.00 & 63.38 & 71.97 & 63.54 & 53.80 \\ 
 & Claude-3.7 & 20.70 & 17.86 & 21.74 & 22.25 & 20.17 & 21.13 & 60.96 & 35.71 & 61.64 & 65.32 & 63.26 & 54.37 \\ 
 & DeepSeek-V3 & 22.20 & 21.43 & 22.22 & 24.57 & 22.93 & 19.15 & 61.99 & 46.42 & 62.42 & 68.79 & 64.92 & 52.39 \\ 
\cmidrule(r){1-2} \cmidrule(r){3-3} \cmidrule(r){4-5} \cmidrule(r){6-8} \cmidrule(r){9-9} \cmidrule(r){10-11} \cmidrule(r){12-14}
 
\multirow{7}{*}{\begin{tabular}[c]{@{}l@{}}\emph{KGQA}\\ \emph{models}\end{tabular}} & EmbedKGQA & 37.16 & 35.71 & 37.20 & 38.73 & 37.57 & 32.39 & 61.05 & 67.86 & 60.87 & 60.98 & 59.94 & 62.25 \\
 & CronKGQA & 42.71 & 39.29 & 42.80 & 43.64 & 46.13 & 35.21 & 64.63 & 67.86 & 64.54 & 64.45 & 66.02 & 63.38 \\
 & TempoQR & 53.06 & 46.43 & 53.24 & 56.36 & 54.14 & 45.07 & 76.11 & 75.00 & 76.04 & 76.01 & 75.96 & 76.05 \\
 & SubGTR & 55.32 & 50.00 & 55.46 & 57.80 & 58.84 & 36.90 & 78.46 & 78.57 & 78.45 & 78.90 & 79.01 & 77.46 \\
 & MultiQA & 51.93 & 46.43 & 52.08 & 55.20 & 57.46 & 39.44 & 76.39 & 78.57 & 76.33 & 77.17 & 75.69 & 76.34 \\ 
 & GeoQA2* & - & - & - & - & - & - & 29.72 & 28.57 & 29.76 & 29.48 & 29.83 & 27.61 \\

\cmidrule(r){2-2} \cmidrule(r){3-3} \cmidrule(r){4-5} \cmidrule(r){6-8} \cmidrule(r){9-9} \cmidrule(r){10-11} \cmidrule(r){12-14}
 
 & STCQA & \textbf{61.52} & \textbf{60.71} & \textbf{61.55} & \textbf{63.01} & \textbf{63.25} & \textbf{53.52} & \textbf{84.29} & \textbf{82.14} & \textbf{84.38} & \textbf{83.82} & \textbf{82.32} & \textbf{86.76} \\ 
\bottomrule
\end{tabular}
}
\caption{Results for the baselines and our method on the STQAD dataset.
GeoQA2* retrieves answers from YAGO2 using SPARQL queries. We apply a lenient criterion: if any of the query results (even if the total number of results significantly exceeds 10) includes a correct answer, we consider the question to be answered correctly. Its result is recorded under the Hits@10 metric.
}
\label{tab5}
\end{table}

While our baseline setting for LLMs is relatively lenient, these models do not perform optimally on spatio-temporal questions. This is because answering such questions necessitates precise spatio-temporal information from the STKG. Despite LLMs possessing extensive world knowledge and billions of parameters, they still struggle to recall precise information, such as geographic coordinates. Our questions frequently require external spatio-temporal factual knowledge, which LLMs cannot handle directly.

A breakdown by constraint type reveals that LLMs and some KGQA
models perform particularly poorly on reasoning tasks involving spatial distances. For instance, GPT-4o's Hits@1 drops by more than 4\% from DDC to DC, indicating significant difficulty in computing or approximating geographic distances without explicit support from the STKG. Similarly, EmbedKGQA lacks spatially-aware embeddings, leading to incorrect entity ranking when directional or distance-based comparisons are required. Despite its rich parametric knowledge, GPT-4o fails to reliably retrieve precise coordinates or timestamps on demand and cannot perform symbolic distance calculations accurately without external tool augmentation. EmbedKGQA, the representative embedding-based baseline, is designed for static KGs and does not model temporal timestamps or geographic coordinates, rendering it ineffective for multi-constraint questions that require joint temporal and spatial filtering. These observations highlight two core challenges in STQAD: (1) extracting implicit spatio-temporal clues from natural language questions, and (2) performing joint reasoning over heterogeneous modalities—semantic, temporal, and spatial—under compositional constraints. We present an example of answers by EmbedKGQA and GPT-4o in~\ref{case_study}.

Introducing spatial or temporal information further improves the performance of STKGQA. 
Compared to EmbedKGQA, designed for traditional KGs, spatial or temporal KGQA methods such as CronKGQA, TempoQR, SubGTR, and GeoQA2 show significant enhancement, indicating that spatio-temporal information is crucial for addressing STKGQA task.
However, these temporal baselines, which only use temporal vectors, neglect spatial information, and their capabilities are significantly affected when answers involve spatial information. In Section~\ref{stkg_embedding_performance}, to further evaluate the reasoning potential of these embedding-based KGQA methods on STKG, we replace their embedding models with our spatial enhanced embedding method. 
Furthermore, although our evaluation criterion is lenient, the spatial KGQA method GeoQA2 performs poorly in answering spatio-temporal questions. It does not consider temporal information, and its template-based query design lacks flexibility when addressing complex questions, making it difficult to comprehend sentence semantics.
The experimental results indicate that, compared to PLMs and LLMs, the STCQA model, which integrates spatio-temporal embeddings, can more effectively acquire precise knowledge and complete reasoning on STKG. Compared to traditional KGQA and temporal KGQA models, STCQA, which can perform spatio-temporal comparisons and distance calculations, excels in reasoning through spatio-temporal clues and constraints within STKGs. Our method significantly surpasses existing benchmarks, establishing itself as a state-of-the-art solution.

\subsection{Ablation Study}
\label{ablation_study}

\paragraph{w/o Entity Annotation} During the Preprocessing module (in Section~\ref{preprocessing_module}), entities in the question are not replaced with special tokens. 

\paragraph{w/o Spatio-temporal Clue Embedding} In the Answer Retrieval module (in Section~\ref{answer_retrieval_module}), explicit timestamp and geographic coordinate embeddings are not used to replace the output of the BERT encoder.

\paragraph{w/o Constraint Filter} In the Answer Filtering module (in Section~\ref{answer_filter_module}), the answers retrieved from the STKG are used directly, without using Constraint Filter to re-verify spatio-temporal clues in the answers. Consequently, no spatio-temporal comparison or distance calculation is conducted.

\paragraph{w/o Temporal Knowledge} We remove all temporal information from the facts in the STKG, meaning that temporal clues and constraints are excluded from both training and inference across the Preprocessing, Answer Retrieval, and Answer Filtering stages.

\paragraph{w/o Spatial Knowledge} We remove all spatial information from the facts in the STKG, such that geographic coordinates and spatial constraints are not utilized during Preprocessing, Answer Retrieval, and Answer Filtering.

\paragraph{w/o Spatio-temporal Knowledge} We remove all spatio-temporal attributes from the STKG facts, effectively disabling the use of both temporal and spatial clues and constraints throughout the entire pipeline.

\begin{table}[htbp]
\centering
  \resizebox{0.8\textwidth}{!}{
  \begin{tabular}{lccc}
    \toprule
     & Hit@1 & Hit@3 & Hit@10\\
    \midrule
    STCQA&\textbf{61.52}&\textbf{76.20}&\textbf{84.29}\\
    w/o Entity Annotation&60.49&74.97&83.07\\
    w/o Spatio-temporal Clue Embedding&59.74&73.75&82.22\\
    w/o Constraint Filter&57.95&71.40&81.28\\
    w/o Temporal Knowledge&37.72&48.17&58.51\\
    w/o Spatial Knowledge&37.54&49.29&59.45\\
    w/o Spatio-temporal Knowledge&29.82&38.19&46.85\\
  \bottomrule
\end{tabular}
}
\label{tab6}
\caption{Ablation study for STCQA.}
\end{table}

Based on Hits@1, the absence of Entity Annotation results in a performance decrease of 1.03\%, indicating that the inherent meaning of entities affects the BERT encoder's understanding of sentence logic semantics. Furthermore, omitting Spatio-temporal Clue Embedding leads to a 1.78\% performance decline, highlighting the necessity of replacing entities with temporal and geographic coordinates in the embedding stage. This approach enables the model to comprehend the spatio-temporal information contained in the questions. Removing the Constraint Filter results in a 3.57\% performance drop in STCQA, demonstrating that embedding methods alone are insufficient for tasks involving the comparison and calculation of spatio-temporal clues, thus underscoring the necessity of this step. These ablation experiments show that processing the implicit spatio-temporal information emphasized in \textbf{Guideline (2)} is crucial to answering questions.

Keeping the input questions unchanged, selectively removing the model’s access to temporal or spatial information in the STKG significantly degrades the performance of STCQA. Under the Hits@1 metric, the complete removal of spatio-temporal information leads to a performance drop of 31.7\%. In the absence of spatio-temporal information, our embedding module reduces to a standard ComplEx, while the Preprocessing and Answer Filtering modules become ineffective. This leads to a substantial drop in performance, particularly on questions requiring temporal or spatial reasoning. These ablation results demonstrate that spatio-temporal knowledge is deeply integrated into all stages of STCQA—embedding, retrieval, and filtering—and highlight the model’s dependence on such information to accurately answer spatio-temporal questions.

\subsection{STKG Embedding Performance}
\label{stkg_embedding_performance}

The embedding-based KGQA models perform poorly on our dataset, which we attribute to their embedding methods' lack of spatial representation of the STKG. Therefore, to evaluate their potential, we replace their embedding models with the embedding method proposed in Section~\ref{answer_retrieval_module}, which is suitable for our STKG. We randomly divide the facts in the STKG into training, validation, and test sets in an 8:1:1 ratio and assess them using the KG embedding task. We employ the scoring functions ComplEx~\cite{trouillon2016complex} and TComplEx~\cite{lacroix2020tensor} used in these KGQA models as baselines for predicting entities. The results of the STKG embeddings, presented in Table~\ref{tab7}, indicate that the embedding method incorporating geographical information demonstrates a significant advantage over existing embedding baselines. \ref{kg_embedding} provides more details of the KG embedding experiment.

\begin{table}[htbp]
\centering
\resizebox{0.45\textwidth}{!}{
  \begin{tabular}{ccccc}
    \toprule
     & Hit@1 & Hit@3 & Hit@10\\
    \midrule
    ComplEx&21.72&34.30&49.96\\
    TComplEx&21.68&34.75&50.15\\
    Ours&\textbf{40.71}&\textbf{48.45}&\textbf{58.41}\\
  \bottomrule
\end{tabular}
}
\caption{Results for STKG embedding.}
\label{tab7}
\end{table}

\begin{figure}[htbp]
  \centering
  \includegraphics[width=\textwidth]{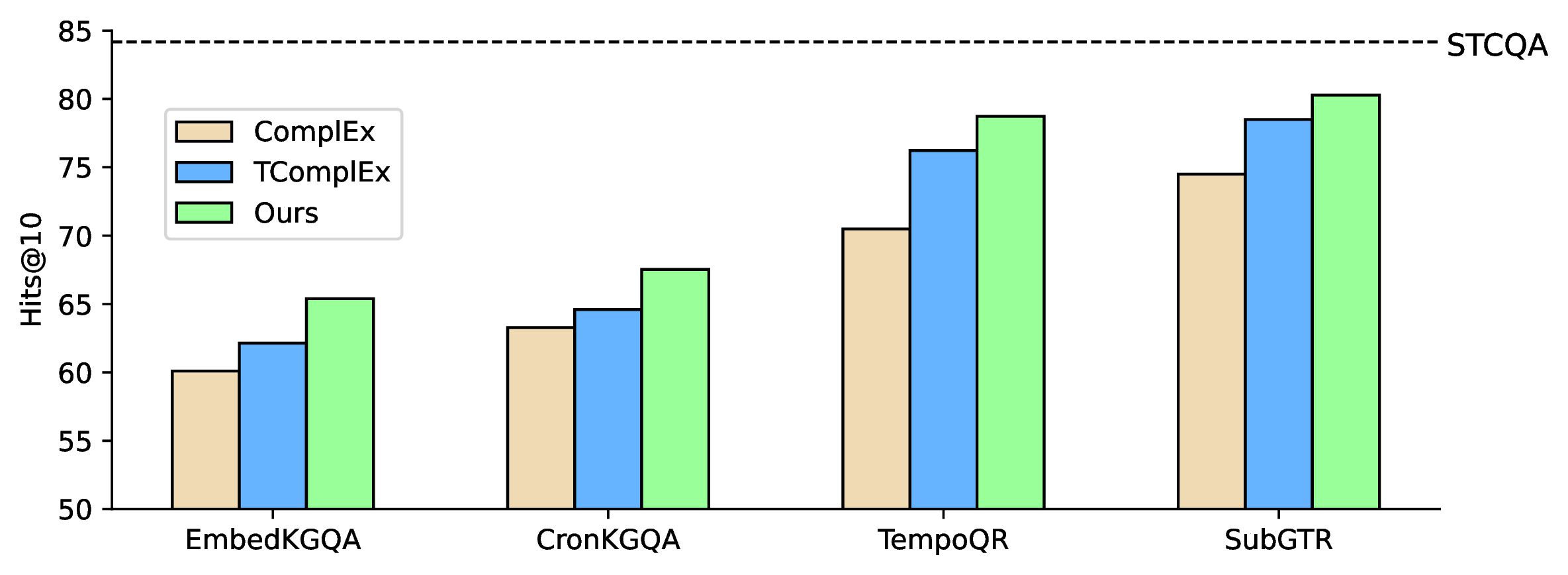}
  \caption{Performance (Hits@10) of embedding-based KGQA baselines on STQAD using different embedding models.}
  \label{fig5}
\end{figure}

Subsequently, we use our embedding method to enhance the KGQA baselines, observing performance improvements in all models. Figure~\ref{fig5} indicates the effectiveness of adding spatial representation for embedding-based KGQA models. 

\subsection{Effect of Spatio-temporal Constraint Relevance}
\label{effect_of_spatio_temporal_constraint_relevance}

\begin{figure}[htpb]
  \centering
  \includegraphics[width=\textwidth]{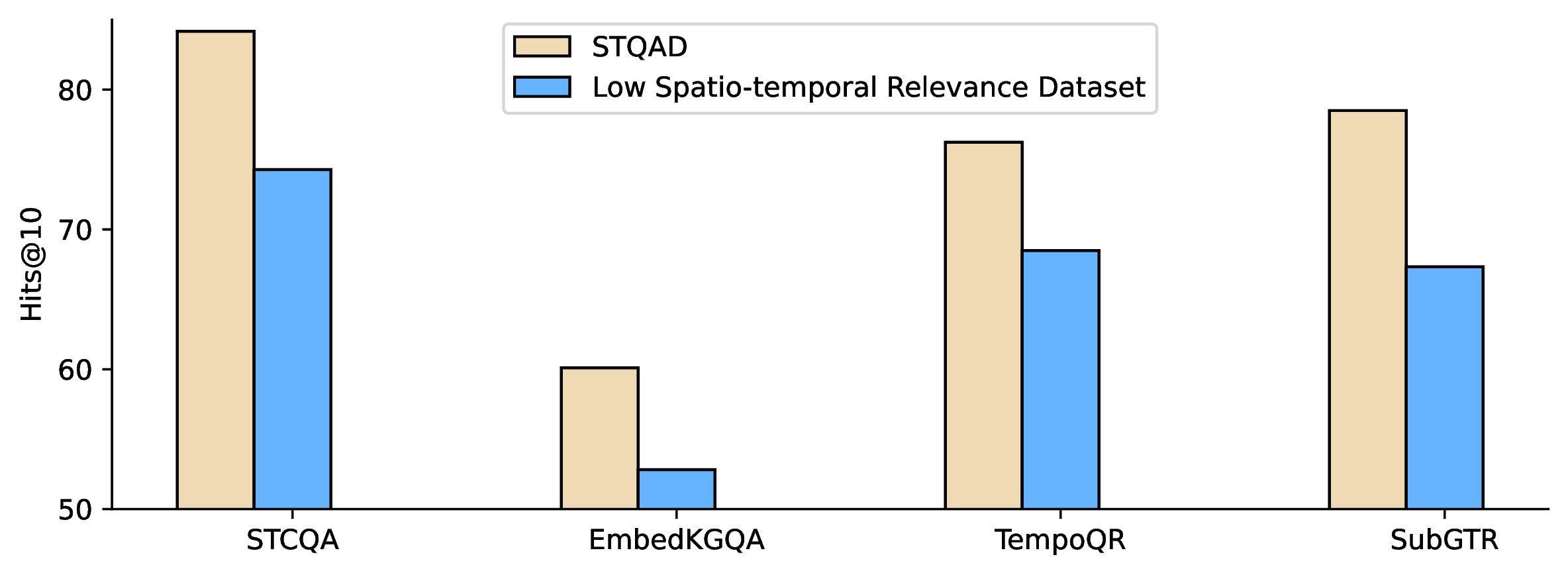}
  \caption{Model performance (Hits@10) on different spatio-temporal constraint relevance dataset.}
  \label{fig6}
\end{figure}

The correlation between spatio-temporal constraints and central facts is a significant indicator for evaluating dataset quality. As \textbf{Guideline(3)} states, using irrelevant entities in the question does not reflect real-world scenarios. We replace the original fact in the question with another fact that satisfies the constraints but is more than 2 hops away from the central fact. This process allowed us to construct a dataset consisting of low co-occurrence entities (such as \emph{Einstein} and \emph{the 25th Academy Awards}). As shown in Figure~\ref{fig6}, we train and evaluate some KGQA models on this dataset, observing a decline in performance. This decline indicates that low-quality data not only fails to meet user requests but also provides limited assistance in enhancing the QA system's performance.

\subsection{Effect of Dataset Size}
\label{effect_of_dataset_size}

\begin{figure}[htpb]
  \centering
  \includegraphics[width=\textwidth]{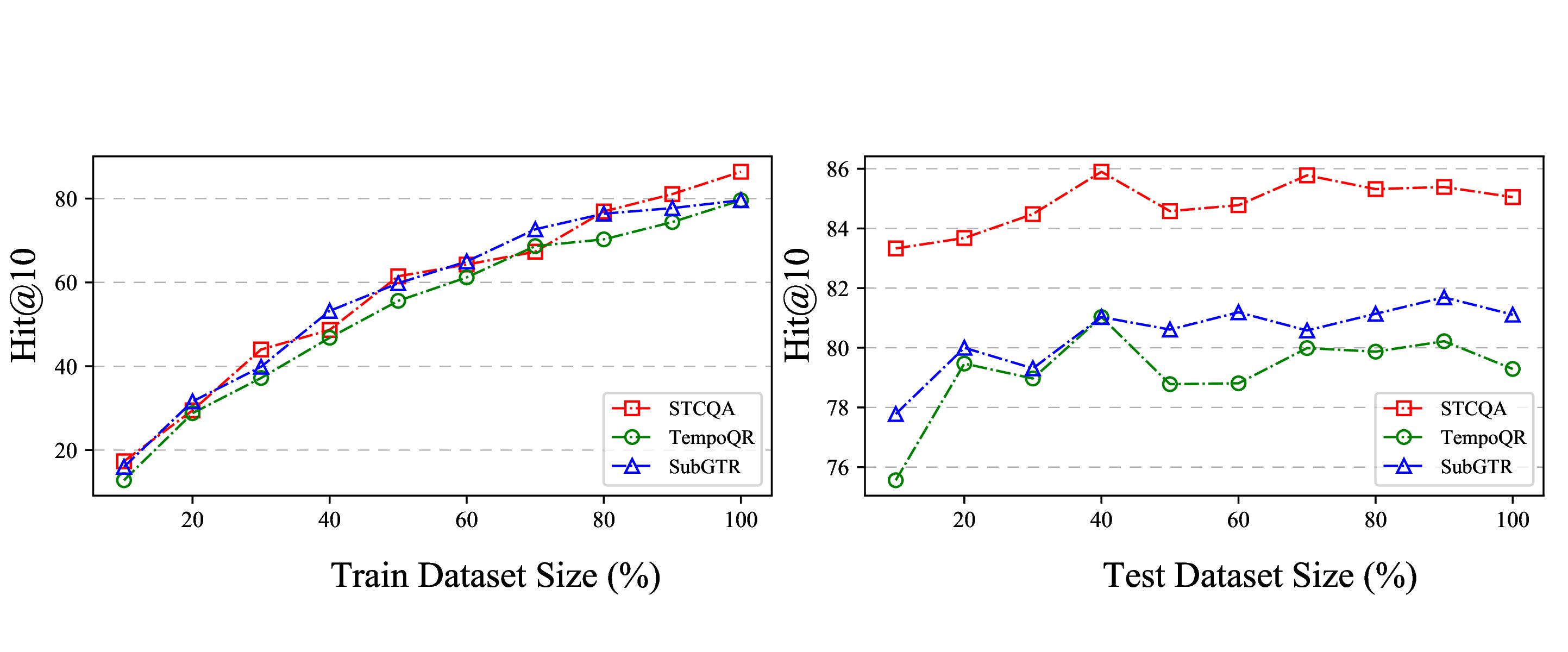}
  \caption{Model performance (Hits@10) vs. dataset size for STCQA, TempoQR and SubGTR.}
  \label{fig7}
\end{figure}

As stated in \textbf{Guideline (4)}, larger datasets typically offer a greater number of training samples, a crucial factor in training accurate and generalizable models. Figure~\ref{fig7} illustrates the impact of dataset size on model performance. Increasing the training dataset size from 10\% to 100\% leads to a steady improvement in model performance. This trend remains consistent across different models, validating the effectiveness of a large-scale dataset for training STKGQA models. 

Similarly, a larger test dataset can yield more reliable and statistically significant evaluation results. With a fixed training dataset size, we progressively enlarge the validation set and test set from 10\% to 100\%. The results obtained from the test dataset demonstrate that employing a larger amount of evaluation data is advantageous for ascertaining a more stable performance across various QA methods implemented on STQAD.

\subsection{The Impact of Knowledge Injection on LLM Performance}
\label{prompt_inject}
Since STCQA explicitly leverages structured knowledge from the STKG, while LLMs are typically provided only with the question text during inference, a potential evaluation imbalance arises. To address this discrepancy and establish a more favorable setting for the LLM, we directly supply it with the 2-hop subgraph extended from the \emph{Central Fact} during data generation and that contains the correct answer, as illustrated in Figure~\ref{fig2}. This approach eliminates the need for the LLMs to perform retrieval over the STKG, thereby isolating their capabilities from the performance of the retriever. We adopt two prompting strategies from prior studies: (1) presenting the subgraph as a sequence of linearized triples ranked by semantic similarity to the question~\cite{Baek2023KnowledgeAugmentedLM}, computed using BERT-based scoring; and (2) first converting the triples into fluent natural language paragraphs using the method in~\cite{wu2023retrieve}, then feeding the text to the LLM. The final prompt template is as follows:

\begin{tcolorbox}[colback=gray!10, colframe=darkgray, title=Knowledge Injection Prompt, width=\textwidth]
{\small
Below are the facts that might be relevant to answer the question, provide the 10 most probable answers in order of likelihood.

Facts: \{facts (linearized triples or natural language paragraphs)\}

Question: \{question\}

Answer:
}
\end{tcolorbox}

As shown in Table \ref{tab8}, explicitly injecting relevant spatio-temporal facts yields substantial gains for all three LLMs. For example, GPT-4o’s Hit@1 increases from 13.73\% to 35.18\% when provided with linearized triples, and further to 41.20\% with natural language paragraphs. The improvement is consistent across Hit@3 and Hit@10, indicating that augmenting LLM prompts with factual STKG context reduces the burden of open-ended retrieval and helps models focus on constraint reasoning. Notably, converting triples into natural language consistently outperforms linearized triples, suggesting that LLMs benefit from semantically fluent and contextually coherent representations of KG facts.

\begin{table}[htbp]
\centering
  \resizebox{0.7\textwidth}{!}{
  \begin{tabular}{lccc}
    \toprule
     & Hit@1 & Hit@3 & Hit@10\\
    \midrule
    ChatGPT & 10.35 & 17.22 & 24.65 \\
    ChatGPT + Linearized Triples & 35.18 & 52.30 & 68.11 \\
    ChatGPT + Natural Language Paragraphs & \textbf{40.07} & \textbf{56.48} & \textbf{72.24} \\
    \midrule
    GPT-4o & 13.73 & 19.29 & 25.21 \\
    GPT-4o + Linearized Triples & 35.74 & 53.63 & 69.45 \\
    GPT-4o + Natural Language Paragraphs & \textbf{41.20} & \textbf{58.14} & \textbf{73.28} \\
    \midrule
    Llama3 & 9.13 & 11.48 & 15.43 \\
    Llama3 + Linearized Triples & 21.35 & 32.17 & 45.34\\
    Llama3 + Natural Language Paragraphs & \textbf{23.05} & \textbf{34.52} & \textbf{47.88} \\
  \bottomrule
\end{tabular}
}
\caption{Results for the LLMs with external knowledge.}
\label{tab8}
\end{table}

However, even with these favorable settings, LLM performance remains well below that of STCQA. The main gap stems from LLMs’ difficulty in executing precise temporal and spatial computations: they must not only judge fact relevance but also reason based on timestamps and geographic distances or orientations under compositional constraints. Without tool integration, such symbolic and numerical reasoning often remains in error. We anticipate that further bridging this gap will require hybrid approaches—equipping LLMs with dedicated reasoning tools (e.g., temporal comparators, spatial calculators) and training them to invoke these tools appropriately. While promising, such tool-augmented pipelines introduce additional API calls and longer response times, making them less efficient than the end-to-end STCQA framework in applications.

\section{Conclusion and Future Work}
In this paper, we introduce STQAD, a new dataset for STKGQA. While there have been numerous KGs based on spatio-temporal information, the existing KGQA datasets lack discussions regarding scenarios involving spatio-temporal information for reasoning. To the best of our knowledge, STQAD is the first dataset that comprises a substantial number of questions that necessitate both temporal and spatial inference. To ensure the realistic situation of the questions in the dataset, we employ strict constraints during question generation. The availability of large datasets enables model evaluation and presents an opportunity for model training. 
Additionally, we propose a novel method, STCQA, serving as a strong baseline for follow-up research.
Due to the limited number of facts in the current STKG, some data imbalances persist, as shown in Table~\ref{tab3}. Nevertheless, we have made a significant first step in this field. In the future, we plan to extend our task to larger-scale STKGs and further design an STKGQA approach based on LLMs.

\section*{Acknowledgments}
This work is supported by the Natural Science Foundation of China (Grant No. U21A20488). We thank the Big Data Computing Center of Southeast University for providing the facility support on the numerical calculations in this paper. 

\appendix
\section{Seed Procedure}
\label{seed_procedure}

\begin{figure}[htbp]
  \centering
    \includegraphics[width=\textwidth]{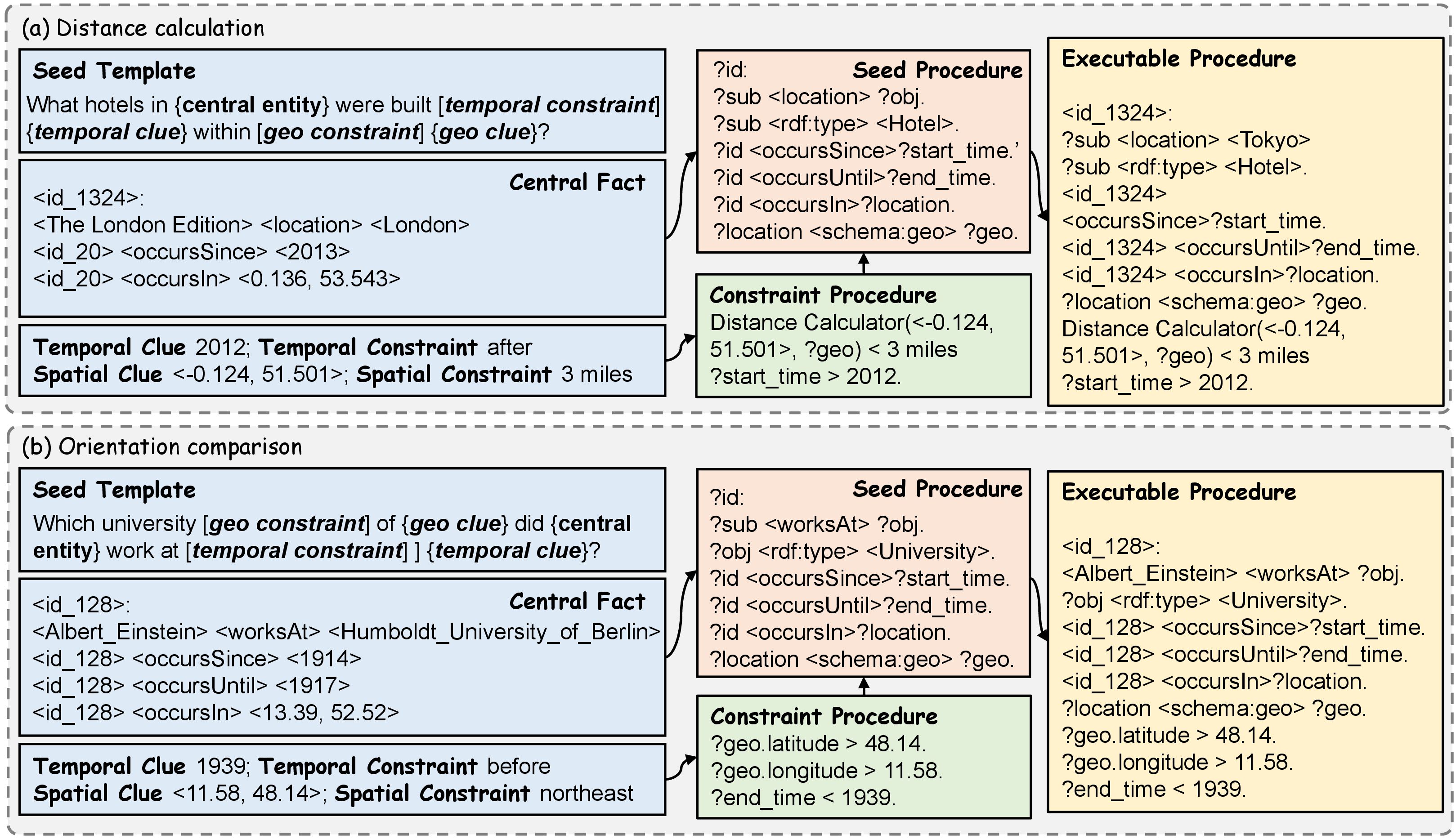}
  \caption{Examples of our seed procedures. Example (a) illustrates distance calculation. We use the \emph{Distance Calculator} based on the Haversine formula to verify geographical constraints. Example (b) demonstrates orientation comparison, where our program can compare all constraints.}
\label{fig8}
\end{figure}

As shown in Figure~\ref{fig8}, each Seed Template corresponds to a Seed Procedure. After obtaining the central fact, spatio-temporal clues, and constraints, we incorporate them into the Seed Procedure to form an Executable Procedure.

The format of the Executable Procedure is defined by YAGO2~\cite{hoffart2013yago2}, and it can be rewritten into SPARQL queries to retrieve answers from the STKG. During the Answer Verification process (in Section~\ref{qa_pair_generation}), this method helps retrieve the correct answers within the STKG. Additionally, by masking the temporal or spatial constraint in the Constraint Procedure, we can observe changes in the number of answers to determine the effectiveness of our spatio-temporal constraints.

\section{Implementation Details}
\label{implementation_details}
For STKG embedding, the embedding dimension D = 512. The model's parameters are updated using Adagrad~\cite{duchi2011adaptive} with a learning rate of 0.1~\cite{lacroix2020tensor}. The training batch size is 1,000. Our embedding method undergoes training for 50 epochs, and the final parameters are determined based on the best validation performance. During STKGQA, the parameters of the pre-trained language model and the STKG embeddings remain unchanged. The layer number of \emph{Transformer} is 2 with 4 heads per layer. The model's parameters are updated using Adam~\cite{kingma2014adam} with a learning rate 2e-5. The training batch size is 150, and the number of epochs is 60. The model updates its parameters by minimizing the cross-entropy loss function, aiming to assign a higher probability to the correct answers.

\section{KG Embedding}
\label{kg_embedding}

ComplEx \cite{trouillon2016complex} represents each entity \(e\) and relation \(r\) as complex vectors \(\mathbf{e}\) and \(\mathbf{r}\), respectively. The score ${\phi}_C$ of a fact $(s, r, o)$ is 
$${\phi}_C\left(\mathbf{e}_s, \mathbf{r}, \overline{\mathbf{e}}_o\right)=\operatorname{Re}\left(\left\langle\mathbf{e}_s,\mathbf{r}, \overline{\mathbf{e}}_o\right\rangle\right)$$
where $\operatorname{Re}(.)$ denotes the real part, $\overline{(.)}$ is the complex conjugate of the embedding vector.
TComplEx \cite{lacroix2020tensor} is an extension of the ComplEx KG embedding method designed for TKGs. TComplEx represents each timestamp $t$ as a complex vector $\mathbf{t}$ and the score  ${\phi}_T$ of a fact $(s, r, o, t)$ is 
$${\phi}_T\left(\mathbf{e}_s, \mathbf{r}, \overline{\mathbf{e}}_o, \mathbf{t}\right)=\operatorname{Re}\left(\left\langle\mathbf{e}_s, \mathbf{r} \odot \mathbf{t}, \overline{\mathbf{e}}_o\right\rangle\right)$$
where $\odot$ is the element-wise product.

Since ComplEx and TComplEx's scoring functions cannot predict timestamps and geographic coordinates, for a fair comparison with our embedding method, we only randomly mask entities in facts to generate train and test data. All embedding methods are trained on the STKG for the completion task.

\section{More Statistics and Detials of STQAD}
\label{more_statistics_and_details_of_stqad}

\begin{figure}[htbp]
    \centering
    \subfloat[Top 10 most occurring entites in questions.]{\includegraphics[width=0.55\textwidth]{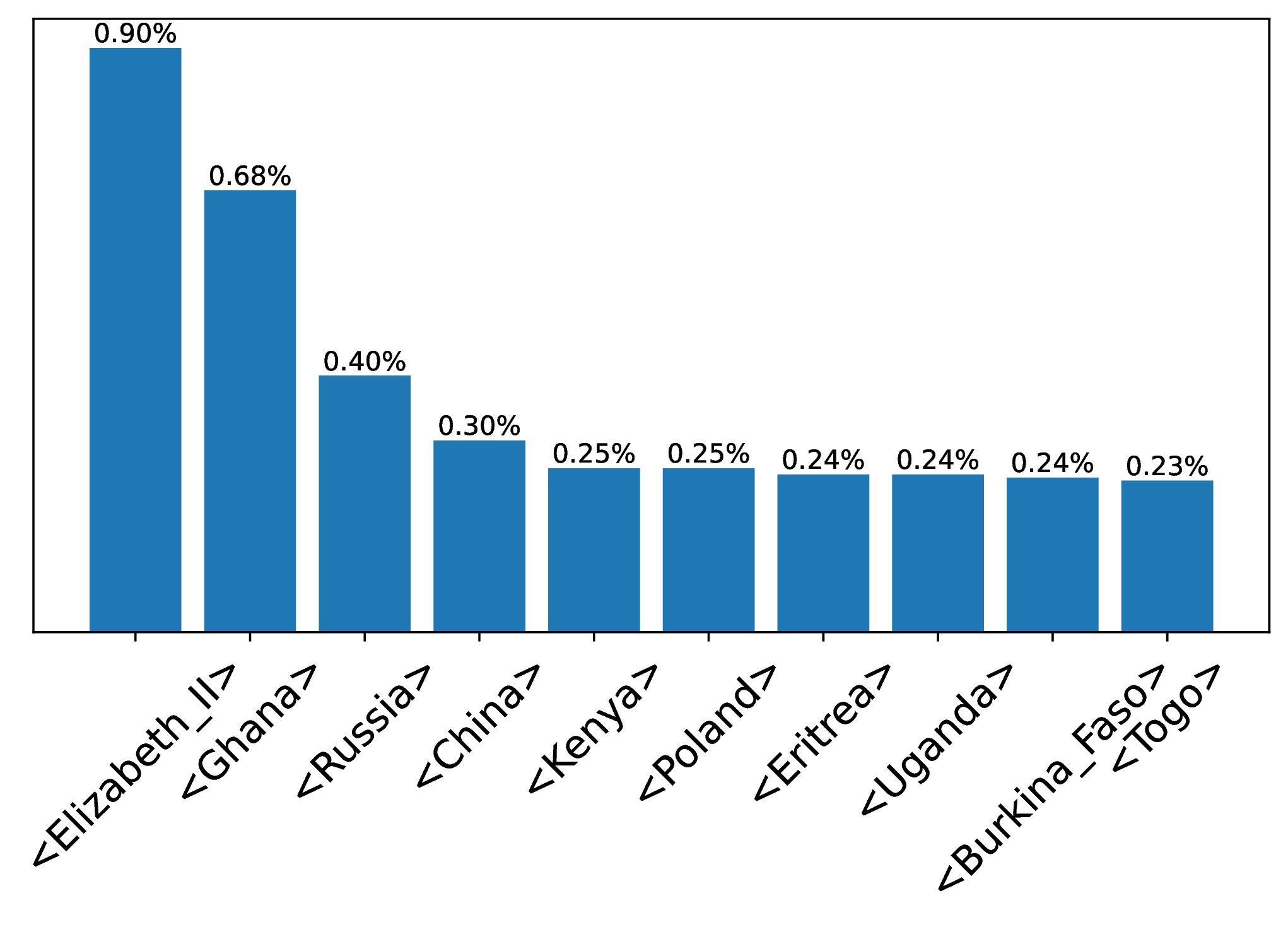}\label{fig9}}
    \hfill
    \subfloat[Distribution of first 3 question words.]{\includegraphics[width=0.45\textwidth]{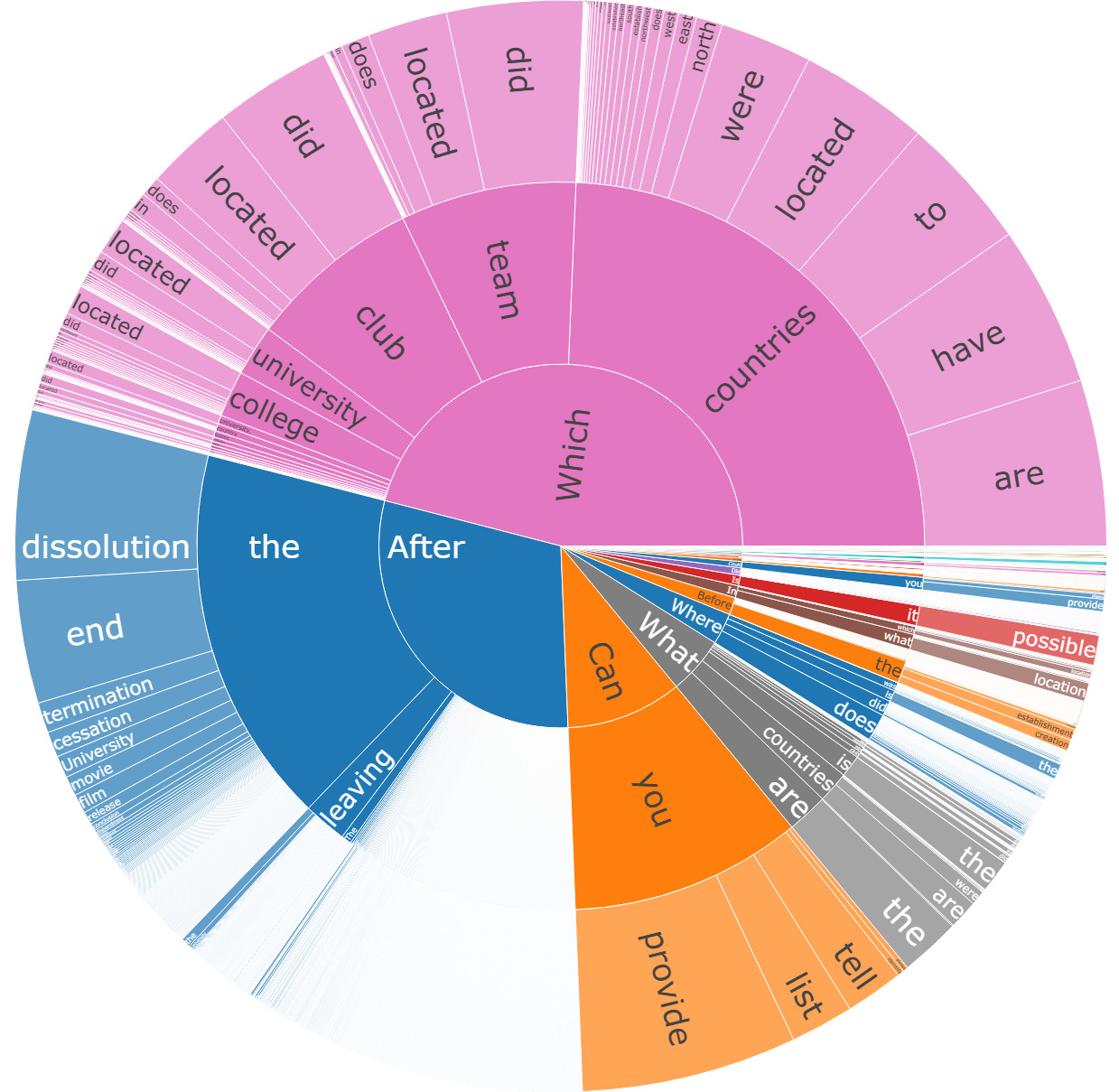}\label{fig10}}
    \caption{More Statistics of STQAD.}
    \label{fig:figure}
\end{figure}

There are 8,092 unique entities in our questions. In Figure~\ref{fig9}, we present the 10 most common entities and their fractions. The distribution of entities within the questions is uniform, indicating the rich representation and no data imbalance in STQAD. 
Figure~\ref{fig10} displays a sunburst of the first 3 words in the questions, showing that questions typically begin with \emph{Which}, \emph{After}, \emph{Can}, \emph{What}, and \emph{Where}. Common topics include \emph{country}, \emph{team}, \emph{club}, and \emph{university}. Additionally, Table~\ref{tab10} provides more examples of questions.

\begin{table}[t]
\resizebox{\columnwidth}{!}{%
\small
\begin{tabular}{>{\centering\arraybackslash}p{2.4cm}p{6.5cm}p{6.3cm}}
\toprule
Relation & \centering\arraybackslash{Template Question} & \centering\arraybackslash{Paraphrased Question} \\ 
  
\cmidrule(r){1-1} \cmidrule(r){2-2} \cmidrule(r){3-3}
  
\textless{}livesIn\textgreater{} &
  \textless{}Akkineni\_Nagarjuna\textgreater{} lives in which place within 310 miles of \textless{}Bengalore\textgreater{} later than the cessation of \textless{}Berlin\_International\_Film\_Festival\textgreater{}? 
  &
  Following the Berlin International Film Festival, where is Akkineni Nagarjuna's residence located, 310 miles away from Bangalore? 
  \\
  
\cmidrule(r){1-1} \cmidrule(r){2-2} \cmidrule(r){3-3}
  
\textless{}memberOf\textgreater{} &
Can you give me the academy that \textless{}Germain\_Bazin\textgreater{} is a member of the northeast of \textless{}Upper\_seine\textgreater{} after leading \textless{}Department\_of\_Paintings\_of\_the\_Louvre\textgreater{}?
  &
Which academy is Germain Bazin a member of located to the northeast of the Upper Seine following his tenure at the Department of Paintings of the Louvre?
  \\
  
\cmidrule(r){1-1} \cmidrule(r){2-2} \cmidrule(r){3-3}
  
\textless{}dealsWith\textgreater{} &
  Before \textless{}Kosovo War\textgreater{}, \textless{}Kosovo\textgreater{} have business deal with which countries southeast of it?
  &
  Prior to the Kosovo War, with which countries to the southeast did Kosovo have business dealings?
  \\

\cmidrule(r){1-1} \cmidrule(r){2-2} \cmidrule(r){3-3}
  
\textless{}isCitizenOf\textgreater{} &
After \textless{}World War II\textgreater{}, \textless{}Paul Schäfer\textgreater{} became a citizen of which country southeast of \textless{}Germany\textgreater{}.
  &
Following World War II, which country to the southeast of Germany did Paul Schäfer become a citizen of?
  \\ 

\cmidrule(r){1-1} \cmidrule(r){2-2} \cmidrule(r){3-3}
  
\textless{}isPoliticianOf\textgreater{} &
  Give me the place that \textless{}George\_W.\_Bush\textgreater{} leads west of \textless{}Oklahoma\textgreater{} before \textless{}the War in Afghanistan\textgreater{}. 
  &
  Before the War in Afghanistan, where did George W. Bush lead west of Oklahoma?
  \\ 

\cmidrule(r){1-1} \cmidrule(r){2-2} \cmidrule(r){3-3}
  
\textless{}graduatedFrom\textgreater{} &
\textless{}Barack\_Obama\textgreater{} be educated by which university within 300 miles of \textless{}New\_York\textgreater{} after working for \textless{}Business\_International\_Corporation\textgreater{}?
  &
 Which university within 300 miles of New York did Barack Obama attend after employment at Business International Corporation?
  \\ 
  \bottomrule
\end{tabular}%
}
\caption{Some representative template questions generated from seed templates.}
\label{tab10}
\end{table}

\section{Case Study}
\label{case_study}

As shown in the examples, GPT-4o fails to recognize the spatio-temporal constraints contained in the question and instead treats the query as ambiguous, refraining from performing reasoning and producing only speculative candidate answers. EmbedKGQA, which cannot model or interpret spatio-temporal constraints, retrieves entities based solely on static relational patterns, returning a list of clubs that the player has been associated with throughout their career, without considering the temporal or spatial conditions specified in the question. These failure cases underscore the limitations of both models in handling constrained spatiotemporal reasoning, whether due to insufficient grounding in spatio-temporal knowledge or the absence of explicit spatio-temporal constraint-aware mechanisms.

\begin{figure}[htbp]
  \centering
  \includegraphics[width=\textwidth]{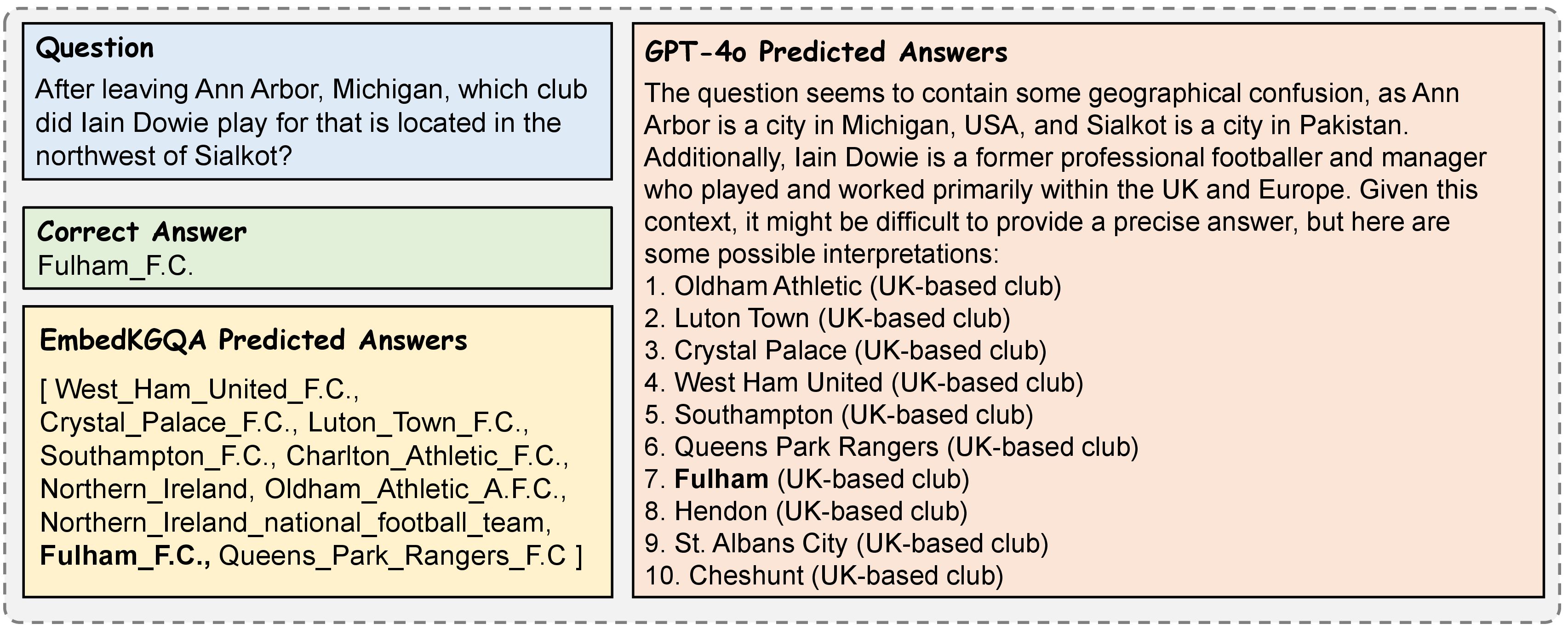}
  \caption{An example of EmbedKGQA and GPT-4o answers.}
  \label{fig11}
\end{figure}

\bibliographystyle{elsarticle-num}
\bibliography{ref}

\end{document}